\documentclass[pdflatex,sn-mathphys-num]{sn-jnl}% Math and Physical Sciences Numbered Reference Style 
\usepackage{graphicx}%
\usepackage{multirow}%
\usepackage{amsmath,amssymb,amsfonts}%
\usepackage{amsthm}%
\usepackage{mathrsfs}%
\usepackage[title]{appendix}%
\usepackage{xcolor}%
\usepackage{textcomp}%
\usepackage{manyfoot}%
\usepackage{booktabs}%
\usepackage{algorithm}%
\usepackage{algorithmicx}%
\usepackage{algpseudocode}%
\usepackage{listings}%
\usepackage{tabularx}
\usepackage{amsmath}
\usepackage{booktabs}
\usepackage{multirow}
\usepackage{caption}
\usepackage{graphicx}
\usepackage{subcaption}
\usepackage{placeins}
\usepackage{comment}
\usepackage{lscape}
\usepackage{color}
\usepackage{colortbl}
\definecolor{electricpurple}{rgb}{0.75, 0.0, 1.0}
 
\newcommand\blue[1]{{\color{blue}#1}} 

\usepackage{adjustbox}

\usepackage{tikz}
\usetikzlibrary{shapes.geometric}

\begin{document}

\title[Article Title]{Enhancing Mass Customization Manufacturing: Multiobjective Metaheuristic Algorithms for flow shop Production in Smart Industry}

%%=============================================================%%
%% GivenName	-> \fnm{Joergen W.}
%% Particle	-> \spfx{van der} -> surname prefix
%% FamilyName	-> \sur{Ploeg}
%% Suffix	-> \sfx{IV}
%% \author*[1,2]{\fnm{Joergen W.} \spfx{van der} \sur{Ploeg} 
%%  \sfx{IV}}\email{iauthor@gmail.com}
%%=============================================================%%

\author*[1]{\fnm{Diego} \sur{Rossit}}\email{diego.rossit@uns.edu.ar}

\author*[1]{\fnm{Daniel} \sur{Rossit}}\email{daniel.rossit@uns.edu.ar}
\equalcont{These authors contributed equally to this work.}

\author*[3]{\fnm{Sergio} \sur{Nesmachnow}}\email{sergion@fing.edu.uy}
\equalcont{These authors contributed equally to this work.}

\affil*[1]{\orgdiv{Engineering Department and INMABB}, \orgname{Universiad Nacional del Sur and CONICET}, \orgaddress{\street{1253 Avenida Alem}, \city{Bahía Blanca}, \postcode{8000}, \state{Buenos Aires}, \country{Argentina}}}

\affil[3]{\orgdiv{Facultad de Ingeniería}, \orgname{Universidad de la República}, \orgaddress{\street{Julio Herrera y Reissig 565}, \city{Montevideo}, \postcode{11300}, \state{Montevideo}, \country{Uruguay}}}

%%==================================%%
%% Sample for unstructured abstract %%
%%==================================%%

\abstract{
The current landscape of massive production industries is undergoing significant transformations driven by emerging customer trends and new smart manufacturing technologies. One such change is the imperative to implement mass customization, wherein products are tailored to individual customer specifications while still ensuring cost efficiency through large-scale production processes. These shifts can profoundly impact various facets of the industry. This study focuses on the necessary adaptations in shop-floor production planning. Specifically, it proposes the use of efficient evolutionary algorithms to tackle the flowshop with missing operations, considering different optimization objectives: makespan, weighted total tardiness, and total completion time. An extensive computational experimentation is conducted across a range of realistic instances, encompassing varying numbers of jobs, operations, and probabilities of missing operations. The findings demonstrate the competitiveness of the proposed approach and enable the identification of the most suitable evolutionary algorithms for addressing this problem. Additionally, the impact of the probability of missing operations on optimization objectives is discussed.
}


%%\pacs[JEL Classification]{D8, H51}

%%\pacs[MSC Classification]{35A01, 65L10, 65L12, 65L20, 65L70}

\maketitle

\section{Introduction}

The concept of the smart industry, often referred to as Industry 4.0, encompasses a transformative paradigm shift in manufacturing and production processes, leveraging advanced technologies such as the Internet of Things, artificial intelligence, robotics, and data analytics~\cite{haverkort2017smart}. It aims to create highly interconnected, data-driven, and adaptive manufacturing ecosystems, enabling real-time monitoring, optimization, and automation of production, supply chains, and services. Smart industry endeavors to enhance efficiency, flexibility, and innovation while fostering sustainable practices, ultimately reshaping traditional industrial practices into agile, intelligent, and interconnected systems.

The smart industry also fosters an interconnected ecosystem where customers, suppliers, and producers collaborate harmoniously~\cite{ibarra2018business}. On the hand of customers, mass customization, a hallmark of this transformation, empowers customers to define their unique product preferences, shaping demand in real-time. On the other hand, through intelligent data-driven systems, suppliers seamlessly adjust their offerings, optimizing inventory and production processes to meet dynamic customization needs. This strategy enhanced interplay among smart industry's technological prowess, empowered customers, and agile suppliers has fundamentally reshaped conventional supply chains through comprehensive integration~\cite{shen2023supply}.

This new paradigm, paves the way to transform the classic production process into mass customization processes, where the client has an active role in the design of the final product. This situation of personalized products, has a significant impact in terms of production processes, since not all the finished product will be the same, then, their production processes must will not be the same. In production systems that are configured as flow shop, this personalization may impact in a missing operation fashion \cite{smutnicki2022cyclic}. In missing operation flow shop scheduling problems, the operation route of each job may be different, where the differences are basically if a job may skip or not one of the operations. Then, the cardinality of the set of operations of jobs is not constant for all jobs. This modification represents a challenging scenario for production scheduling decision-making, because the orders to be planned are not all the same \cite{rossit2021solving}. Furthermore, decision makers must fulfill many criteria for solving the scheduling of production efficiently nowadays, then, the complexity and difficulty of the problem enhances. 

This article addresses a missing operation, multi-objective, flow shop scheduling problem using a metaheuristic approach~\cite{nesmachnow2014overview}. Mainly, the problem considered is a regular flow shop system, where there is one machine or production resource per stage, and the jobs to be processed by that system may not require to be processed in every machine. Also, as mentioned before, to optimize this problem involves to consider simultaneously more than one criterion. In this case three different objective functions are analyzed, namely, makespan, total tardiness and total completion time. These goals treated as a multi-objective optimization problem, enable to optimize production system utilization, customer service level and production orders flow, respectively. As far as the authors know, this is the first time that a missing operation flow shop problem with three objectives is studied. The metaheuristics applied to solve the problem are NSGA-II, NSGA-III, MOEA/D, SPEA2, and AGEMOEA-II. 

This article is an extension of our previous conference paper entitled ``Smart Industry Strategies for Shop-Floor Production Planning Problems to Support Mass Customization'' presented at the VII Ibero-American Congress of Smart Cities that was held in Mexico City and Cuernavaca in November, 2023~\cite{rossit2023smart}. The new content includes further description of the conceptual problem with a realistic illustrative example, inclusion of another state-of-the-art multi-objective evolutionary algorithm to address the target problem, an extended computational experimentation over a larger set of instances with larger percentage of missing operations, and an extended analysis of the new results. Thus, the main contributions of this work include: a novel description of the relation between the flowshop scheduling problem with missing operations and mass customization process of the Industry 4.0 environments including an illustrative example from a real industrial process; a suit of state-of-the-art multi-objective metaheuristics algorithms to address the flowshop scheduling problem with missing operations considering as optimization objectives the makespan, total tardiness and total completion time; and an extensive computational experimentation to asses the efficiency of the different metaheuristics to solve this problem.

The article is structured as follows. Section~\ref{sec_flowshop} formally presents the flowshop problem, describing its mathematical formulation and the main related works. Section~\ref{sec_resolution} presents the metaheuristic algorithms used for the resolution of the multi-objective flowshop problem. Section~\ref{sec_experiment} describes the computational experimentation, including the implementation details, the description of instances and the main results. Finally, Section~\ref{sec_conclusion} presents the conclusions of the research and formulates the main lines for future work.

\section{Mass customization and the multi-objective flowshop problem with missing operations\label{sec_flowshop}}

This section presents a comprehensive presentation of mass customization in Smart Industry environments. Then, a detailed description of the problem addressed in this work is introduced, where the objectives function considered for the multi-objective approach are mathematically described. Finally, the related works found in literature are revised in order to highlight the main contributions of the reported research.

\subsection{Mass customization impact on the shop-floor operations}

Smart industry enables manufacturers to gather insights from customer preferences, adapt production processes, optimize resource allocation, and dynamically reconfigure assembly lines, resulting in the cost-effective creation of highly customized products on a scale previously unattainable. 
Mass customization starts with an intelligent smart product design, in which the preferences of the user are translated into instructions for the shop-floor operations on how to plan the production phase. 
As aforementioned, an important aspect of smart industry is mass customization. Mass customization refers to the capacity of efficiently producing goods and services that are tailored to meet individual customer preferences and requirements, while still achieving economies of scale similar to mass production~\cite{baranauskas2020mapping}. Several companies have successfully invested in enhancing their mass customization strategies to offer personalized products to their customers. One example is
Nike, which allows customers to design their own sneakers through its Nike By You  platform, where they can choose colors, materials, and customize various design elements~\cite{nikeExample}. Another example is Dell, which offers customized computers and laptops through its website, allowing customers to select hardware specifications, software options, and even design their own laptop skins~\cite{dellExample}. Larger products also have entered to this wave of customization. For example in cars production, BMW enables customers to personalize their luxury vehicles with a wide range of custom features, including paint colors, interior materials, and technology options~\cite{BMWExample}. Similarly, Tesla allows customers to personalize various aspects of their electric vehicles, including battery range, interior features, and autopilot capabilities~\cite{teslaExample}. 

In this regard, mass customization has a huge impact in shop-floor operations~\cite{zawadzki2016smart}. Among the aspects that are involved in an efficient shop-floor management are: 
i) Workflow flexibility: shop floors must be designed to accommodate varying product configurations and customization options. Flexible layouts, modular workstations, and adaptable production lines are essential to seamlessly transition between different customization processes~\cite{keddis2015modeling}. In this sense, quick changeover processes, e.g., Single-Minute Exchange of Die (SMED), are crucial to transition between different customization options efficiently minimizing downtime during changes.
ii) Real-time data integration: mass customization relies on real-time data from various sources, such as customer orders, inventory levels, and production progress. Shop-floor operations need to integrate data systems to ensure accurate and up-to-date information for decision-making. Technology based on RFID devices is an ally in this sense~\cite{zhong2013rfid}. 
iii) Agile production planning: traditional batch production planning may not align with the requirements of mass customization. Agile production planning methods, such as lean manufacturing or just-in-time principles, become more relevant to optimize resources and minimize waste~\cite{ding2023combining};
iv) Inventory management: mass customization demands a delicate balance between maintaining a diverse inventory of components and minimizing excess stock. Inventory management systems must be optimized to ensure that the right components are available for each customization option~\cite{guo2018inventory};
v) Advanced manufacturing technologies: Mass customization often leverages advanced technologies like robotics, automation, and additive manufacturing. These technologies enhance efficiency, precision, and speed in producing customized products on the shop floor~\cite{savolainen2020additive};
vi) Quality 4.0: the varying processes that emerge with diverse customization options, enforce quality control to adapt to ensure consistent quality across all product variations. Thus, robust quality assurance protocols and testing procedures are crucial to maintain customer satisfaction~\cite{zonnenshain2020quality};
vii) Skilled Workforce: Shop-floor operators need to be skilled in handling various customization processes, technologies, and equipment. Training programs are essential to empower workers with the knowledge needed to execute customization tasks effectively~\cite{liboni2019smart};
viii) Smart production planning and control: aims to intelligently perform the activities of loading, scheduling, sequencing, monitoring, and controlling the use of resources and materials during production by means of data analytics, AI, and machine learning~\cite{oluyisola2022designing}. 

\begin{figure}[h]
\centering
\includegraphics[width=1.0\textwidth]{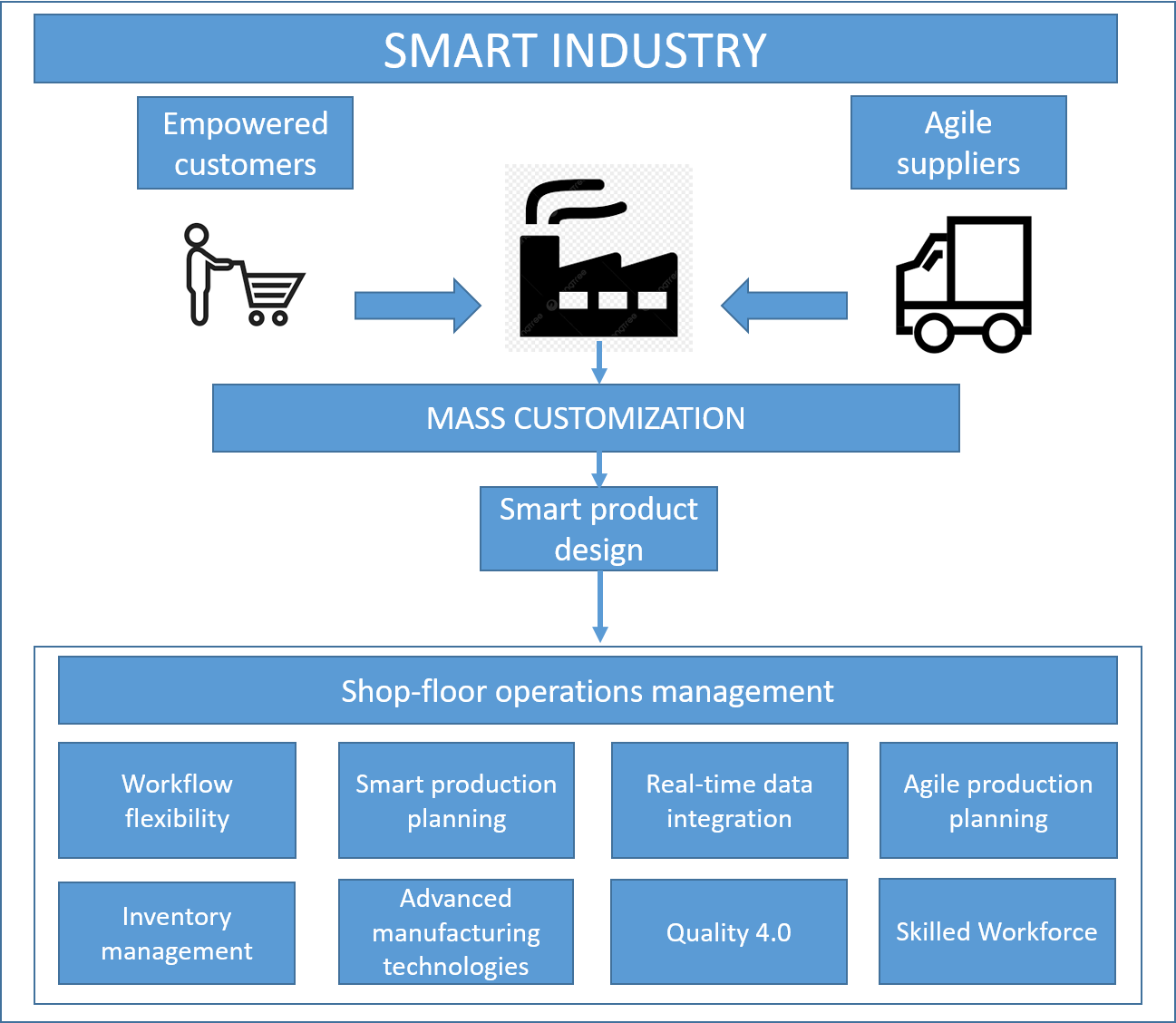}
\caption{Smart industry: the new paradigm and the impact on shop-floor operations management.} 
\label{fig:smart_industry}
\end{figure}

Fig.~\ref{fig:smart_industry} presents a summary of the main concepts involved in an Smart industry and the impact of mass customization to shop-floor operations management. This article focuses on smart production planning proposing new resolution methodologies to solve the flowshop problem that arises in the context of mass customization with missing operations.

%\iffalse
%\fi
%\vspace{-0.2cm}
%
%
\subsection{An illustrative example to describe the relation between mass customization and missing operations}

\begin{figure}
\includegraphics[width=\textwidth]{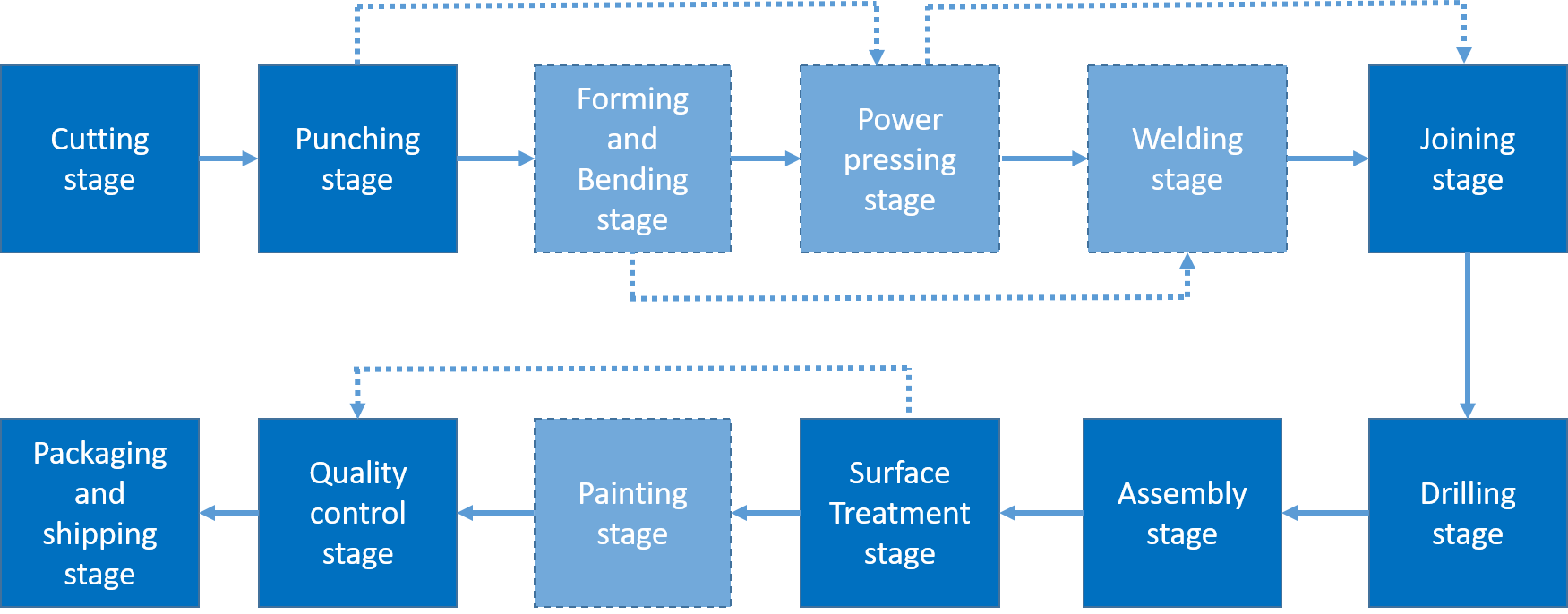}
\caption{Example of the industrial process of steel furniture with missing operations. \label{fig:example_missing}}
\end{figure}

To analyze the relation between mass customization and missing operations let's consider as an example the industrial process of the steel furniture as is depicted in Fig.~\ref{fig:example_missing}. This production process have several stages. However, not all the stages have to be used for every product. One example is the stage of forming and pressing. The standar production pace would demand a "forming and bending stage" after the "punching stage", and after this, a power pressing stage comes, where the final plastic deformations are achieved. However, some of this furniture may skip "forming and bending stage", since it is not necessary to that piece of furniture. The same goes for "power pressing stage", this type of deformation is very precise and time consuming, then, it can be not necessary in all furniture. Finally, the painting stage can be missed since several customers can decide to have their furniture in steel original colour.

%\iffalse
%\fi

The flowshop problem is a classic NP-hard optimization problem in manufacturing that deals with sequencing operations on machines to minimize makespan, total completion time, total tardiness or another desired objective linked to the production efficiency~\cite{rossit2018non}. When considering mass customization, the flowshop problem with missing operations adds complexity to the optimization process~\cite{rossit2021flow}. 
%As aforementioned, missing operations refer to situations where not all operations are required for every product variant, leading to variations in processing sequences. 
Solving the flowshop problem with missing operations requires designing production schedules that consider the varying processing sequences for different product configurations. Thus, advanced scheduling algorithms and optimization techniques need to be employed to create adaptable production planning schedules that can efficiently accommodate the customization requirements of a diverse product range while optimizing the desired optimization criteria, ultimately contributing to the successful implementation of mass customization strategies.

\subsection{Mathematical formulation}

The mathematical formulation of the multi-objective flowshop problem with missing operations considers the following elements:
\begin{itemize}
    \item A set of machines or operations $M$ that can be performed.
    \item A set of jobs $J$ that have to be delivered.
    \item A due date $d_{j}$ in which each job has to be delivered.
    \item Given the matrix $P_{JM}$ which contains the processing times $p_{jm}$ for each job $j$ on each machine $m$ and the processing order of jobs on the machines $Or$, a completion time function $C(j): P_{JM} \times Or \rightarrow \mathcal{R}_{0}^{+}$ that returns the completion time of job $j$, i.e., the time when the job has performed all the required operations in all the machines.
    \item A vector $w_j$ that gives the relative importance of each job regarding the total completion time.
\end{itemize}

Then, the optimization problem addressed in this work is to define the processing order of jobs on the machines that simultaneously  minimized  the following three metrics: the makespan, the weighted total completion time and the total tardiness, which are computed as in Eqs.~\eqref{Eq:makespan}-\eqref{Eq:total_tardiness}.

\begin{subequations}
\begin{alignat}{3}
\mbox{min} \quad & \text{makespan} = \displaystyle \max_{j\in{J}} \{C(j)\} \label{Eq:makespan}\\ 
\mbox{min} \quad & \text{Total Completion Time} = \displaystyle\sum_{j\in{J}} w_j \times C(j) \label{Eq:total_completation_Time}\\
\mbox{min} \quad & \text{Total Tardiness} = \displaystyle\sum_{j\in{J}} max(0;C(j) - d_j) \label{Eq:total_tardiness}
\end{alignat}
\end{subequations}

The impact of missing operations affects the matrix of processing times $P_{JM}$ since several parameters $p_{jm}$ can be equal to 0. Regarding constraints, the problem at hand is bound by specific restrictions. First, there are non-overlapping constraints in place, which prohibit more than one job from being processed on the same machine simultaneously. Second, there are logical timely constraints, meaning that the start time for processing a job on a machine must occur after the finish time of the same job on the previous machine. These restrictions shape the flowshop problem which is known to be a computationally complex problem~\cite{garey1976complexity}.

\subsection{\blue{Literature Review}}
Flow shop problems with missing operation have been studied for many years \cite{glass1999two}. These problems allow us to represent different variants of production processes where physical assets allow a range of opportunities greater than those requested by each specific product. However, since the digital revolution of production processes, the interest in this type of problems has grown considerably, allowing the incorporation of customization or particular requirements that some clients demand \cite{dios2018eficientes,aheleroff2020digital}. As for instance, furniture industry where the different pieces of furniture request a subset of the operations available in the shop-floor, then the scheduling problem can be approached as a missing operation flow shop problem. This problem was addressed in \cite{saravanan2014optimization}, where the authors tackled a stainless furniture manufacturing process, and considered makespan as objective function. the optimization procedure was based on Simulated Annealing. Then, in \cite{saravanan2016minimization} the same problem was revisited, but in this case the objective function studied was total tardiness, and the authors used a Genetic algorithm to optimize it. Other intresting case, is at concrete industry sector, the production process is configured as a flow shop with missing operation and customer requests personalized goods. In Smutnicki et al.~\cite{smutnicki2022cyclic}, the objective is to  minimize the cycle time, and for this, the authors developed a comprehensive innovative approach that overcomes the particular restrictions the production process has, like lags between operations. Other type of problem is tackled at Ramezanian et al.~\cite{ramezanian2017milp}, where a non-permutation flow shop problem with missing operation is considered. Mathematical programming models are developed to optimize the makespan in this work.  \blue{Another non-permutation flow shop problem is approached at Rossit et al.~\cite{rossit2021solving}. The objective function considered is total tardiness and a two-step meta-heuristic algorithm is developed. In \cite{marichelvam2014performance} a scatter search algorithm is developed to optimize the makespan a missing operation hybrid flow shop}. More recently in Han et al.~\cite{han2022scheduling} a missing operation flow shop problem arises at the semiconductor industry. At this case, a special feature of the studied problem is the presence of time waiting constraints, and the objective function is the minimization of total tardiness. For solving the problem heuristic algorithm are implemented. \blue{An interesting case appears in the pharmaceutical industry, where the physical-chemical analysis processes can have some missing operation depending on the type of final product. This problem was addressed by \cite{de2022novel}, where they consider the total tardiness as objective function and used a two-step procedure, firstly, they used a matheuristics where some variables are fixed in order to give priority to some jobs in the schedule, and then, secondly a genetic algorithm optimize the rest of the variables.}
Regarding multi-objective optimization with missing operation in regular flow shop problems, the literature is more scarce. Basically, as far as the authors know, it can be found only in Rossit et al.~\cite{rossit2021explicit,rossit2022multi}, where in these works a bi-objective problem is approached by means of Evolutionary algorithms. In those studies makespan and total tardiness are minimized, and different levels of missing operations are considered.

\section{Resolution approach}
\label{sec_resolution}

This section describes the proposed resolution approach for solving the considered multi-objective flowshop problem.

\subsection{Overall description and algorithms}

Various strategies have been proposed in the literature to solve multi-objective optimization problems. Resolution approaches include exact methods rooted in mathematical programming~\cite{rossit2020exact}, as well as heuristic and metaheuristic strategies~\cite{nesmachnow2018comparison}. For complex combinatorial challenges like the one described in this paper, metaheuristics offer an efficient resolution strategy to attain high-quality solutions in reasonable computing times and, thus, have been extensively used in similar problems~\cite{rossit2021solving,rossit2021explicit}.

Among metaheuristics, multi-objective evolutionary algorithm (MOEAs) are population-based methods inspired by the evolutionary process of species in nature. MOEAs have demonstrated to be successful methods with application in diverse complex optimization problems~\cite{nesmachnow2014overview}. Particularly, this article proposes applying five state-of-the-art evolutionary metaheuristics to address the target problem: SPEA2, NSGA-II, NSGA-III, MOEA/D, and AGEMOEA-II, which are described next.

\subsubsection{Strength Pareto Evolutionary Algorithm 2 (SPEA2)} 
SPEA2 is a MOEA that focuses on non-dominated sorting and density estimation to generate a diverse set of solutions, allowing for effective exploration of the Pareto front. In this line, a notable aspect of SPEA2 is its fitness calculation, which takes into account both Pareto dominance and diversity. The algorithm introduces the concept of strength to gauge how many candidate solutions are dominated by or dominate other solutions. Additionally, fitness assignment involves density estimation. Elitism is also incorporated through the use of a population that stores non-dominated individuals discovered during the search. 

\subsubsection{Non-dominated Sorting Genetic Algorithm II (NSGA-II)} 
NSGA-II is a widely used evolutionary algorithm that employs non-dominated sorting, crowding distance, and elitism to evolve a diverse population of solutions, efficiently approximating the Pareto front. NSGA-II is characterized by an evolutionary search using a non-dominated elitist ordering that diminishes the complexity of the dominance check, a crowding technique for diversity preservation, and a fitness assignment method considering dominance ranks and crowding distance values. All these features are integrated to provide a robust and effective search, which has been successfully applied to solve multi-objective optimization problems in many application areas. 

\subsubsection{Non-dominated Sorting Genetic Algorithm III (NSGA-III)} 
NSGA-III is an extension of NSGA-II that incorporates reference points to guide the optimization process, enhancing the spread of solutions along the Pareto front and supporting better convergence. 
\subsubsection{Multi-objective Evolutionary Algorithm based on Decomposition (MOEA/D)} \sloppy MOEA/D decomposes a multi-objective optimization problem into subproblems, each solved by a separate optimization process. It balances exploration and exploitation to efficiently approximate the Pareto front by iteratively updating solutions through collaboration among subproblems.

\subsubsection{AGE-MOEA: Adaptive Geometry Estimation based MOEA II (AGEMOEA-II)} \sloppy AGEMOEA-II is recently new version of the previous AGEMOEA metaheuristic~\cite{panichella2022improved}. The algorithm follows a similar structure to NSGA-II but incorporates a modified crowding distance formula. The evolutionary process begins by sorting non-dominated fronts through a non-dominated sorting procedure. Subsequently, the initial front serves for objective space normalization and estimation of Pareto front geometry. Then, the Minkowski p-norm is then employed to calculate a survival score using as a reference point the closest solution from the middle of the first front. The survival score amalgamates both distance from neighbors and proximity to the ideal point, i.e., the unattainable point that is obtained by combining the best values achieved for each objective function in any solution of the front.
Then, each individual is initially compared based on rank, followed by evaluation using the survival score, which encapsulates both proximity and dispersion.

\subsection{Description of the proposed metaheuristics}

The proposed MOEAs operate using the following features:

%\vspace{-1mm}
\subsubsection{Solution representation} As it usual in similar works, solutions are denoted by permutations of integers within a vector. The index placement within the vector represents the processing sequence on the initial machine, with the associated integer values corresponding to individual jobs slated for scheduling. Thus, the length of the vector represents to the total job count.

%\vspace{-1mm}
\subsubsection{Initialization} The population, comprising $\#P$ individuals, is initialized through a random procedure that generates permutations devoid of repeated integer values. Employing a uniform probability distribution, each value within a solution representation is chosen from the interval [1,n].

%\vspace{-1mm}
\subsubsection{Evolutionary operators} 
The well-known Partially Mapped Crossover (PMX) is employed as the recombination operator. This crossing mechanism pairs two chosen individuals with a probability of $p_c$, and it has been widely utilized in various studies tackling permutation-encoded scheduling issues. Subsequently, the mutation operator relies on Swap Mutation, involving the interchange of two elements within the permutation. Application of the mutation operator to an individual occurs with a probability of $p_m$. Notably, the proposed operators ensure the feasibility of the resultant solutions.

\section{Computational experimentation\label{sec_experiment}}

This section presents the computational experimentation of the proposed approach, including the description of instances, the methodology used for the experimental evaluation, and the main numerical results.

\subsection{Description of the problem instances} 

A set of realistic instances were constructed for the computational experimentation, following the procedure 
by Henneberg and Neufeld
\cite{henneberg2016constructive}. Processing times were generated as integer values within the range [0:100] following a pseudo-uniform distribution, with the probability of a processing time been zero with a relatively higher value compared to the other possible processing times. 
This approach ensured the existence of varied processing times including the possibility of missing operations. The sets of instances were constructed considering three different numbers of jobs (30, 40 and 50), two different numbers of machines or operations (10 and 20) and five different percentage probability of missing operations (0\%, 10\%, 20\%, 40\%, and 60\%), these instances where designed following the recommendations given in other studies of the literature, such as \cite{dios2018efficient,han2022scheduling} . The instances were named using the following convention $n$/$m$/$p$/$i$, where $n$ is the number of jobs, $m$ is the number of machines, $p$ for the percentage probability of missing operations, and $i$ for the identification number to differentiate instances that coincide in number of jobs, number of machines and percentage probability of missing operations.

\subsection{Methodology for the computational experimentation}

This subsection presents the description of how the computational experimentation of the proposed MOEAs is performed.

\subsubsection{Implementation details and execution platform} The implementation of the proposed MOEAs was carried out in Java, using the JMetal framework version 6.1~\cite{nebro2021evolving}.
The computational experimentation phase was executed on the National Supercomputing Center, Uruguay (Cluster-UY)~\cite{nesmachnow2019cluster}.

\subsubsection{Evaluation metrics} The evaluation is performed considering two 
multi-objective optimization metrics: spread and relative hypervolume (RHV). 
Spread~\cite{deb2001} is a metric of diversity that evaluates the distribution of the non-dominated solutions, assessing the capacity of correctly sampling the Pareto front. Unlike other typical distribution metrics such as spacing, the spread as formulated in Eq.~\eqref{eq:spread} takes into account the information about the extreme points of the true Pareto front to calculate a more accurate value of the dispersion. 

\begin{equation}
    \text{spread} = \frac{\sum_{o\in \mathcal{O}} d_{o}^{e} + \sum_{i\in \mathcal{ND}} |\overline{d}-d_i|}{\sum_{o\in \mathcal{O}} d_{o}^{e} + |\mathcal{ND}| \overline{d}}
    \label{eq:spread}
\end{equation}
\vspace{3mm}

In Eq.~\eqref{eq:spread}, $\mathcal{O}$ is the set of objectives, $\mathcal{ND}$ is the set of non-dominated solutions, $d_{o}^{e}$ is the distance between the extreme point of the Pareto front regarding objective $o$ and the closest non-dominated solution in the computed Pareto front, $d_i$ is the distance between the non-dominated solution $i$ in the computed
Pareto front and the closest neighbor non-dominated solution, and $\overline{d}$ is the average value of all $d_i$. On the other hand, the RHV quantifies the ratio between the hypervolumes (in the search space of the objective functions) covered by the computed Pareto front and the true Pareto front of the problem. Thus, in an ideal situation the RHV value equals one. Consequently, RHV serves as a comprehensive metric that evaluates both numerical accuracy (proximity of the computed Pareto front to the real Pareto front) and the distribution of the non-dominated solutions. When the true Pareto front is unknown for a problem instance, as it is the case in this study, the true Pareto front is approximated using all the non-dominated solutions obtained from all the resolutions performed for that instance.

\subsubsection{Parameterization} 
The determination of the optimal parametric configuration was guided by statistical analysis. This process was pivotal in establishing the values for the key parameters of the studied MOEAs: population size ($\#P$), crossover probability ($p_c$), and mutation probability ($p_m$). To determine these parameters different values were assessed: 50 and 100 for population size, 0.5, 0.7, and 0.9 crossover probabilities, and 0.01, 0.05, and 0.1 mutation probabilities. Consequently, a comprehensive evaluation encompassing sixteen parametric configurations ensued for each of the five MOEAs. The analysis for the parameter setting was based on the RHV, which as aforementioned is a robust summary metric. The stopping condition was set to 150,000 evaluations of the objective function. For the comparison three small instances different from the main computational study were used.

As the RHV values did not follow a normal distribution according to the Shapiro-Wilk test, the Friedman rank test, a non-parametric method, was employed to assess the goodness of each configuration. Particularly, the neighborhood size of the MOEA/D was chosen in 3\% of $\#P$ which showed a good performance in our previous work~\cite{rossit2022multi}. After the parameter setting, the following configurations were chosen for the studied MOEAs:
\begin{itemize}
    \item AGEMOEA-II: $\#P = 100$, $p_c = 0.7$, and $p_m=0.1$
    \item MOEA/D: $\#P = 50$, $p_c = 0.5$, and $p_m=0.1$
    \item NSGA-II: $\#P = 100$, $p_c = 0.7$, and $p_m=0.1$
    \item NSGA-III: $\#P = 50$, $p_c = 0.7$, and $p_m=0.1$
    \item SPEA2: $\#P = 100$, $p_c = 0.9$, and $p_m=0.1$
\end{itemize}

\subsection{Numerical results} 

This subsection describes the result of the computational experimentation. For each instance and each MOEA, 30 independent executions were performed.

\subsubsection{Multi-objective optimization metrics} 
The summary of the results of the studied multi-objective optimization metrics are reported in Table~\ref{tab:rhv} (RHV metric) and Table~\ref{tab:spread} (spread metric).
The tables report the statistical test used to study if there are significant differences among the medians or averages, a central tendency and a dispersion measure for the studied MOEAs. In the instances in which results follow a normal distribution, the ANOVA test is applied as statistical test (expressed with ``A'' in the Tables~\ref{tab:rhv} and~\ref{tab:spread}) and the mean and standard deviation are used as central tendency and dispersion measures respectively. Conversely, in the case of non-parametric distributions, Kruskal-Wallis (expressed with ``KW'' in the table) is applied as statistical test, and the median and  interquartile range are used as central tendency and dispersion measures respectively. For each instance, the best result is 
marked with bold font. Results with colour background (blue or gray) indicate the best mean or median obtained for each instance. In particular, results marked with blue background indicate the cases in which the test verified a significant statistical difference with respect to the other MOEAs. Conversely, the gray background indicate that the test did not verified statistical difference between the best MOEA and at least one of the other MOEAs.

Regarding RHV, NSGA-II obtained the largest
mean/median in 20 out of 60 instances. SPEA2 obtained the largest mean/median in 19 out of 60 instances. NSGA-III obtained the largest mean/median in 13 out of 60 instances. Finally, AGEMOEA-II, obtained the largest mean/median in 8 out of 60 instances. The largest mean/median value was obtained by SPEA2 for instances 30/20/10/2 (0.8081). 

In terms of spread, 
SPEA2 obtained the smallest value in 40 out of 60 instances. NSGA-II and MOEA/D obtained the smallest values in 12 out of 60 instances and in 8 out of 60 instances, respectively.
The overall smallest value of spread was obtained by SPEA2 for instance 30/20/10/1 (0.3967). Overall the SPEA2 and the NSGA-II had the best performance for the instances studied, been able to outperformed the other MOEAs in both analyzed metrics.

%\iffalse
\begin{table}[!htbp]
\setlength{\tabcolsep}{2pt}
\renewcommand{\arraystretch}{0.96}
\adjustbox{max width=\textwidth}{%
\centering
\begin{tabular}{ll rr r rr r rr r rr r rr}
\toprule
& & \multicolumn{2}{c}{AGEMOEA} & & \multicolumn{2}{c}{MOEA/D} & & \multicolumn{2}{c}{NSGA-II} & & \multicolumn{2}{c}{NSGA-III} & & \multicolumn{2}{c}{SPEA2} \\ 
\cmidrule{3-4}\cmidrule{6-7}\cmidrule{9-10}\cmidrule{12-13}\cmidrule{15-16}
Instance & Test & \begin{tabular}{c}mean/\\median\end{tabular} & \begin{tabular}{c}std/\\iqr\end{tabular} & & \begin{tabular}{c}mean/\\median\end{tabular} & \begin{tabular}{c}std/\\iqr\end{tabular} & & \begin{tabular}{c}mean/\\median\end{tabular} & \begin{tabular}{c}std/\\iqr\end{tabular} & & \begin{tabular}{c}mean/\\median\end{tabular} & \begin{tabular}{c}std/\\iqr\end{tabular} & & \begin{tabular}{c}mean/\\median\end{tabular} & \begin{tabular}{c}std/\\iqr\end{tabular} \\ 
\midrule
30/10/0/1 & KW & 0.7004 & 0.0712 &  & 0.4758 & 0.0936 &  & 0.7045 & 0.0692 &  & 0.7384 & 0.1007 &  & \cellcolor[HTML]{BDD7EE}0.7646 & 0.0526 \\
30/10/0/2 & KW & \cellcolor[HTML]{BDD7EE}0.6752 & 0.1477 &  & 0.3951 & 0.1022 &  & 0.6556 & 0.1204 &  & 0.6720 & 0.0884 &  & 0.6704 & 0.0876 \\
30/20/0/1 & KW & 0.7207 & 0.0757 &  & 0.4986 & 0.0706 &  & 0.6867 & 0.0874 &  & 0.7012 & 0.0679 &  & \cellcolor[HTML]{BDD7EE}0.7260 & 0.0630 \\
30/20/0/2 & KW & \cellcolor[HTML]{BDD7EE}0.7820 & 0.0545 &  & 0.5170 & 0.0996 &  & 0.7793 & 0.0580 &  & 0.7565 & 0.0540 &  & 0.7773 & 0.0615 \\
40/10/0/1 & A & 0.6009 & 0.1182 &  & 0.4090 & 0.1051 &  & 0.6181 & 0.1176 &  & 0.6105 & 0.1326 &  & \cellcolor[HTML]{D0CECE}0.6668 & 0.0907 \\
40/10/0/2 & A & 0.6151 & 0.1367 &  & 0.4601 & 0.1384 &  & 0.6642 & 0.1137 &  & 0.6815 & 0.1106 &  & \cellcolor[HTML]{D0CECE}0.6880 & 0.1105 \\
40/20/0/1 & A & \cellcolor[HTML]{D0CECE}0.7329 & 0.0605 &  & 0.4806 & 0.0758 &  & 0.7268 & 0.0681 &  & 0.7284 & 0.0706 &  & 0.7291 & 0.0712 \\
40/20/0/2 & KW & 0.5665 & 0.1265 &  & 0.3628 & 0.1202 &  & 0.5649 & 0.1566 &  & \cellcolor[HTML]{BDD7EE}0.5874 & 0.1146 &  & 0.5672 & 0.1603 \\
50/10/0/1 & A & 0.4848 & 0.1494 &  & 0.4041 & 0.1359 &  & \cellcolor[HTML]{BDD7EE}0.5683 & 0.1229 &  & 0.5420 & 0.1264 &  & 0.5320 & 0.1250 \\
50/10/0/2 & A & 0.5152 & 0.1331 &  & 0.3666 & 0.0940 &  & 0.6004 & 0.1190 &  & 0.5961 & 0.1096 &  & \cellcolor[HTML]{D0CECE}0.6506 & 0.0916 \\
50/20/0/1 & KW & 0.6201 & 0.1299 &  & 0.4384 & 0.1064 &  & 0.6153 & 0.0732 &  & \cellcolor[HTML]{BDD7EE}0.6551 & 0.1222 &  & 0.6421 & 0.1392 \\
50/20/0/2 & A & 0.6549 & 0.0713 &  & 0.4551 & 0.0993 &  & 0.6274 & 0.0874 &  & \cellcolor[HTML]{D0CECE}0.6624 & 0.0948 &  & 0.6513 & 0.0890 \\
30/10/10/1 & A & 0.6548 & 0.1276 &  & 0.5897 & 0.0992 &  & 0.7139 & 0.1186 &  & \cellcolor[HTML]{D0CECE}0.7150 & 0.1099 &  & 0.7133 & 0.1005 \\
30/10/10/2 & A & 0.6468 & 0.0817 &  & 0.4686 & 0.1237 &  & 0.6786 & 0.0934 &  & \cellcolor[HTML]{D0CECE}0.6931 & 0.1167 &  & 0.6756 & 0.0966 \\
30/20/10/1 & A & 0.7661 & 0.0472 &  & 0.5693 & 0.0733 &  & \cellcolor[HTML]{D0CECE}0.7701 & 0.0611 &  & 0.7514 & 0.0653 &  & 0.7601 & 0.0516 \\
30/20/10/2 & KW & 0.7752 & 0.0798 &  & 0.5871 & 0.0808 &  & 0.8070 & 0.0563 &  & 0.7901 & 0.0662 &  & \cellcolor[HTML]{BDD7EE}0.8081 & 0.0351 \\
40/10/10/1 & A & 0.4452 & 0.1514 &  & 0.2786 & 0.1591 &  & \cellcolor[HTML]{D0CECE}0.5825 & 0.1649 &  & 0.5237 & 0.1550 &  & 0.5454 & 0.1520 \\
40/10/10/2 & A & 0.4920 & 0.1625 &  & 0.3967 & 0.1180 &  & 0.6281 & 0.1166 &  & \cellcolor[HTML]{D0CECE}0.6479 & 0.1174 &  & 0.5709 & 0.1585 \\
40/20/10/1 & A & \cellcolor[HTML]{D0CECE}0.6516 & 0.0860 &  & 0.4160 & 0.0867 &  & 0.6331 & 0.0931 &  & 0.6335 & 0.1235 &  & 0.6060 & 0.1144 \\
40/20/10/2 & KW & 0.6665 & 0.1083 &  & 0.4332 & 0.0630 &  & \cellcolor[HTML]{BDD7EE}0.6799 & 0.1339 &  & 0.6753 & 0.0807 &  & 0.6647 & 0.1314 \\
50/10/10/1 & A & 0.4742 & 0.1220 &  & 0.3345 & 0.1258 &  & 0.5453 & 0.1253 &  & \cellcolor[HTML]{D0CECE}0.5789 & 0.1137 &  & 0.5291 & 0.1119 \\
50/10/10/2 & KW & 0.5408 & 0.1819 &  & 0.3936 & 0.1416 &  & 0.6407 & 0.1348 &  & 0.5974 & 0.1799 &  & \cellcolor[HTML]{BDD7EE}0.6831 & 0.1490 \\
50/20/10/1 & A & 0.6468 & 0.0953 &  & 0.3901 & 0.1077 &  & 0.6401 & 0.1210 &  & \cellcolor[HTML]{D0CECE}0.6588 & 0.1311 &  & 0.6583 & 0.1004 \\
50/20/10/2 & KW & 0.6352 & 0.1442 &  & 0.4293 & 0.1819 &  & 0.6542 & 0.1579 &  & \cellcolor[HTML]{BDD7EE}0.6563 & 0.1387 &  & 0.6391 & 0.1267 \\
30/10/20/1 & KW & 0.6256 & 0.1900 &  & 0.4244 & 0.0615 &  & 0.7410 & 0.1269 &  & 0.6298 & 0.1128 &  & \cellcolor[HTML]{BDD7EE}0.7549 & 0.1177 \\
30/10/20/2 & KW & 0.7366 & 0.1238 &  & 0.4700 & 0.1346 &  & \cellcolor[HTML]{BDD7EE}0.7703 & 0.1124 &  & 0.6947 & 0.0945 &  & 0.7671 & 0.0894 \\
30/20/20/1 & KW & 0.6934 & 0.0713 &  & 0.4531 & 0.0421 &  & \cellcolor[HTML]{BDD7EE}0.7114 & 0.0888 &  & 0.7016 & 0.0650 &  & 0.6658 & 0.0826 \\
30/20/20/2 & KW & 0.7167 & 0.0761 &  & 0.5074 & 0.0772 &  & 0.7412 & 0.0371 &  & 0.7452 & 0.0644 &  & \cellcolor[HTML]{BDD7EE}0.7548 & 0.0643 \\
40/10/20/1 & A & 0.5361 & 0.1080 &  & 0.4267 & 0.1257 &  & 0.6030 & 0.1174 &  & 0.5929 & 0.1348 &  & \cellcolor[HTML]{D0CECE}0.6422 & 0.1297 \\
40/10/20/2 & A & 0.3945 & 0.1448 &  & 0.3284 & 0.1230 &  & \cellcolor[HTML]{D0CECE}0.5680 & 0.1387 &  & 0.5311 & 0.1132 &  & 0.5120 & 0.1160 \\
40/20/20/1 & KW & \cellcolor[HTML]{BDD7EE}0.5920 & 0.2184 &  & 0.3709 & 0.1248 &  & 0.5808 & 0.3007 &  & 0.5192 & 0.1520 &  & 0.5463 & 0.0935 \\
40/20/20/2 & A & 0.6733 & 0.0961 &  & 0.4383 & 0.0937 &  & 0.6779 & 0.1068 &  & 0.6648 & 0.1108 &  & \cellcolor[HTML]{D0CECE}0.6937 & 0.1060 \\
50/10/20/1 & KW & 0.5950 & 0.1163 &  & 0.4461 & 0.1466 &  & 0.7054 & 0.0704 &  & 0.5893 & 0.1706 &  & \cellcolor[HTML]{BDD7EE}0.7230 & 0.1491 \\
50/10/20/2 & KW & 0.4898 & 0.1304 &  & 0.4046 & 0.1501 &  & \cellcolor[HTML]{D0CECE}0.6193 & 0.1333 &  & 0.6013 & 0.1493 &  & 0.5976 & 0.1184 \\
50/20/20/1 & A & 0.5703 & 0.1443 &  & 0.3956 & 0.1220 &  & 0.4886 & 0.1264 &  & \cellcolor[HTML]{D0CECE}0.5999 & 0.1297 &  & 0.5205 & 0.1331 \\
50/20/20/2 & A & 0.5909 & 0.0975 &  & 0.4435 & 0.1074 &  & \cellcolor[HTML]{D0CECE}0.6531 & 0.1276 &  & 0.6239 & 0.1194 &  & 0.6442 & 0.1050 \\
30/10/40/1 & A & 0.7174 & 0.0914 &  & 0.4970 & 0.0671 &  & \cellcolor[HTML]{D0CECE}0.7557 & 0.0749 &  & 0.6830 & 0.0807 &  & 0.7420 & 0.0614 \\
30/10/40/2 & KW & 0.6345 & 0.1047 &  & 0.4588 & 0.0977 &  & 0.6415 & 0.0951 &  & 0.6145 & 0.1502 &  & \cellcolor[HTML]{BDD7EE}0.6674 & 0.0986 \\
30/20/40/1 & A & 0.7448 & 0.0554 &  & 0.5627 & 0.0662 &  & \cellcolor[HTML]{D0CECE}0.7478 & 0.0843 &  & 0.6949 & 0.0725 &  & 0.7383 & 0.0923 \\
30/20/40/2 & A & \cellcolor[HTML]{D0CECE}0.7141 & 0.0630 &  & 0.4670 & 0.0613 &  & 0.7107 & 0.0764 &  & 0.6859 & 0.0902 &  & 0.6975 & 0.0836 \\
40/10/40/1 & KW & 0.6837 & 0.1395 &  & 0.5834 & 0.1709 &  & \cellcolor[HTML]{BDD7EE}0.7817 & 0.1064 &  & 0.7081 & 0.1395 &  & 0.7630 & 0.1117 \\
40/10/40/2 & A & 0.5392 & 0.1431 &  & 0.4775 & 0.1536 &  & 0.6293 & 0.1466 &  & 0.6226 & 0.1741 &  & \cellcolor[HTML]{D0CECE}0.6505 & 0.1266 \\
40/20/40/1 & A & 0.5951 & 0.0971 &  & 0.3979 & 0.1088 &  & 0.6288 & 0.1033 &  & 0.6298 & 0.0846 &  & \cellcolor[HTML]{D0CECE}0.6483 & 0.1140 \\
40/20/40/2 & A & \cellcolor[HTML]{D0CECE}0.6405 & 0.1114 &  & 0.4468 & 0.0896 &  & 0.6010 & 0.0902 &  & 0.6301 & 0.0897 &  & 0.6307 & 0.0831 \\
50/10/40/1 & A & 0.5192 & 0.1218 &  & 0.4118 & 0.1119 &  & \cellcolor[HTML]{D0CECE}0.6386 & 0.1327 &  & 0.5627 & 0.1295 &  & 0.5907 & 0.0912 \\
50/10/40/2 & A & 0.5229 & 0.1031 &  & 0.4752 & 0.1247 &  & 0.6089 & 0.1139 &  & \cellcolor[HTML]{D0CECE}0.6271 & 0.1080 &  & 0.6234 & 0.1018 \\
50/20/40/1 & KW & 0.5644 & 0.1331 &  & 0.3679 & 0.1317 &  & \cellcolor[HTML]{BDD7EE}0.6306 & 0.0883 &  & 0.5496 & 0.1682 &  & 0.5869 & 0.0971 \\
50/20/40/2 & A & 0.5255 & 0.1184 &  & 0.3714 & 0.0785 &  & \cellcolor[HTML]{D0CECE}0.5914 & 0.0889 &  & 0.5673 & 0.1134 &  & 0.5433 & 0.0950 \\
30/10/60/1 & KW & 0.7043 & 0.0694 &  & 0.5571 & 0.0950 &  & \cellcolor[HTML]{BDD7EE}0.7790 & 0.0662 &  & 0.7151 & 0.0692 &  & 0.7710 & 0.0787 \\
30/10/60/2 & KW & 0.6873 & 0.0974 &  & 0.5186 & 0.1466 &  & 0.7516 & 0.1060 &  & 0.7167 & 0.0988 &  & \cellcolor[HTML]{BDD7EE}0.7527 & 0.1090 \\
30/20/60/1 & A & 0.7465 & 0.0571 &  & 0.5777 & 0.0744 &  & \cellcolor[HTML]{D0CECE}0.7873 & 0.0562 &  & 0.7424 & 0.0763 &  & 0.7732 & 0.0695 \\
30/20/60/2 & A & \cellcolor[HTML]{D0CECE}0.7439 & 0.0472 &  & 0.5482 & 0.0562 &  & 0.7428 & 0.0558 &  & 0.7182 & 0.0628 &  & 0.7418 & 0.0720 \\
40/10/60/1 & KW & 0.5273 & 0.2222 &  & 0.3472 & 0.1410 &  & \cellcolor[HTML]{BDD7EE}0.6015 & 0.1359 &  & 0.5755 & 0.1262 &  & 0.5999 & 0.2118 \\
40/10/60/2 & A & 0.5957 & 0.1039 &  & 0.4349 & 0.0859 &  & 0.6560 & 0.0683 &  & 0.6248 & 0.1006 &  & \cellcolor[HTML]{D0CECE}0.7010 & 0.1106 \\
40/20/60/1 & A & 0.5859 & 0.1134 &  & 0.3882 & 0.0946 &  & \cellcolor[HTML]{D0CECE}0.6411 & 0.1011 &  & 0.6012 & 0.1105 &  & 0.6322 & 0.0979 \\
40/20/60/2 & A & 0.5356 & 0.1173 &  & 0.4154 & 0.1217 &  & \cellcolor[HTML]{D0CECE}0.5942 & 0.1238 &  & 0.5708 & 0.1235 &  & 0.5565 & 0.1163 \\
50/10/60/1 & A & 0.4031 & 0.1058 &  & 0.4163 & 0.1170 &  & 0.4804 & 0.0988 &  & \cellcolor[HTML]{BDD7EE}0.5820 & 0.1204 &  & 0.4956 & 0.1234 \\
50/10/60/2 & KW & 0.6090 & 0.1588 &  & 0.5563 & 0.1594 &  & 0.6962 & 0.0803 &  & 0.6775 & 0.1022 &  & \cellcolor[HTML]{BDD7EE}0.7516 & 0.1297 \\
50/20/60/1 & A & 0.5333 & 0.0945 &  & 0.3830 & 0.0668 &  & 0.5758 & 0.1031 &  & \cellcolor[HTML]{D0CECE}0.6018 & 0.1043 &  & 0.5939 & 0.0707 \\
50/20/60/2 & A & 0.4829 & 0.1102 &  & 0.3654 & 0.1093 &  & 0.5235 & 0.0915 &  & 0.5314 & 0.1286 &  & \cellcolor[HTML]{D0CECE}0.5600 & 0.1337 \\
\bottomrule
\end{tabular}}
\captionof{table}{Results of RHV metric for the studied MOEAs.\label{tab:rhv}}
\end{table}

\begin{table}[!htbp]
\setlength{\tabcolsep}{2pt}
\renewcommand{\arraystretch}{0.96}
\adjustbox{max width=\textwidth}{%
\centering
\begin{tabular}{ll rr r rr r rr r rr r rr}
\toprule
& & \multicolumn{2}{c}{AGEMOEA} & & \multicolumn{2}{c}{MOEA/D} & & \multicolumn{2}{c}{NSGA-II} & & \multicolumn{2}{c}{NSGA-III} & & \multicolumn{2}{c}{SPEA2} \\ 
\cmidrule{3-4}\cmidrule{6-7}\cmidrule{9-10}\cmidrule{12-13}\cmidrule{15-16}
Instance & Test & \begin{tabular}{c}mean/\\median\end{tabular} & \begin{tabular}{c}std/\\iqr\end{tabular} & & \begin{tabular}{c}mean/\\median\end{tabular} & \begin{tabular}{c}std/\\iqr\end{tabular} & & \begin{tabular}{c}mean/\\median\end{tabular} & \begin{tabular}{c}std/\\iqr\end{tabular} & & \begin{tabular}{c}mean/\\median\end{tabular} & \begin{tabular}{c}std/\\iqr\end{tabular} & & \begin{tabular}{c}mean/\\median\end{tabular} & \begin{tabular}{c}std/\\iqr\end{tabular} \\ 
\midrule
30/10/0/1 & A & 0.6777 & 0.1175 &  & 0.6552 & 0.2019 &  & 0.5401 & 0.0505 &  & 0.5484 & 0.0722 &  & \cellcolor[HTML]{BDD7EE}0.4320 & 0.0555 \\
30/10/0/2 & KW & 0.6245 & 0.1310 &  & 0.7248 & 0.2785 &  & 0.5212 & 0.0808 &  & 0.5142 & 0.0920 &  & \cellcolor[HTML]{D0CECE}0.4021 & 0.0796 \\
30/20/0/1 & KW & 0.6537 & 0.1693 &  & \cellcolor[HTML]{BDD7EE}0.4876 & 0.1223 &  & 0.5798 & 0.0912 &  & 0.5565 & 0.0627 &  & 0.4889 & 0.0736 \\
30/20/0/2 & KW & 0.6382 & 0.1286 &  & \cellcolor[HTML]{BDD7EE}0.4678 & 0.1967 &  & 0.5262 & 0.0684 &  & 0.5422 & 0.0820 &  & 0.4705 & 0.0880 \\
40/10/0/1 & A & 0.8286 & 0.0864 &  & 0.8821 & 0.2341 &  & 0.6893 & 0.1245 &  & 0.7840 & 0.1160 &  & \cellcolor[HTML]{D0CECE}0.6834 & 0.0969 \\
40/10/0/2 & KW & 0.7762 & 0.1663 &  & 0.8359 & 0.3967 &  & \cellcolor[HTML]{D0CECE}0.6615 & 0.1488 &  & 0.6623 & 0.1490 &  & 0.6903 & 0.1646 \\
40/20/0/1 & KW & 0.6622 & 0.1373 &  & \cellcolor[HTML]{BDD7EE}0.4464 & 0.2817 &  & 0.5578 & 0.0851 &  & 0.5993 & 0.0941 &  & 0.5186 & 0.0759 \\
40/20/0/2 & KW & 0.6018 & 0.0978 &  & \cellcolor[HTML]{BDD7EE}0.5056 & 0.2425 &  & 0.6182 & 0.0728 &  & 0.5859 & 0.0947 &  & 0.5668 & 0.1114 \\
50/10/0/1 & KW & 0.9287 & 0.0910 &  & 1.0042 & 0.3843 &  & 0.8414 & 0.1455 &  & 0.9476 & 0.1080 &  & \cellcolor[HTML]{D0CECE}0.8175 & 0.1870 \\
50/10/0/2 & A & 0.8987 & 0.1078 &  & 0.9179 & 0.2717 &  & 0.8164 & 0.1017 &  & 0.8315 & 0.1252 &  & \cellcolor[HTML]{BDD7EE}0.7512 & 0.1313 \\
50/20/0/1 & KW & 0.6575 & 0.1455 &  & 0.7095 & 0.4181 &  & 0.5722 & 0.0629 &  & 0.5835 & 0.1177 &  & \cellcolor[HTML]{D0CECE}0.5433 & 0.0820 \\
50/20/0/2 & A & 0.7007 & 0.0997 &  & 0.7712 & 0.2345 &  & 0.6115 & 0.0610 &  & 0.6703 & 0.1154 &  & \cellcolor[HTML]{D0CECE}0.5773 & 0.0849 \\
30/10/10/1 & A & 0.7119 & 0.0849 &  & 0.7175 & 0.1994 &  & 0.6578 & 0.0770 &  & 0.6815 & 0.0830 &  & \cellcolor[HTML]{D0CECE}0.6208 & 0.0744 \\
30/10/10/2 & A & 0.7193 & 0.0791 &  & 0.6519 & 0.2026 &  & \cellcolor[HTML]{D0CECE}0.6476 & 0.0878 &  & 0.7052 & 0.0736 &  & 0.6770 & 0.0658 \\
30/20/10/1 & KW & 0.5915 & 0.0837 &  & 0.5162 & 0.2997 &  & 0.5264 & 0.0599 &  & 0.5703 & 0.0661 &  & \cellcolor[HTML]{D0CECE}0.3967 & 0.0902 \\
30/20/10/2 & KW & 0.6876 & 0.0784 &  & 0.5994 & 0.2166 &  & 0.5955 & 0.0862 &  & 0.6123 & 0.0627 &  & \cellcolor[HTML]{BDD7EE}0.5208 & 0.0492 \\
40/10/10/1 & KW & 0.9219 & 0.0919 &  & 1.0194 & 0.4078 &  & 0.8853 & 0.1059 &  & 0.9362 & 0.0995 &  & \cellcolor[HTML]{D0CECE}0.7452 & 0.1865 \\
40/10/10/2 & KW & 0.7791 & 0.1128 &  & 0.9046 & 0.4606 &  & 0.7506 & 0.1277 &  & 0.7526 & 0.1608 &  & \cellcolor[HTML]{D0CECE}0.7122 & 0.1360 \\
40/20/10/1 & KW & 0.6514 & 0.1088 &  & 0.6437 & 0.4194 &  & 0.5704 & 0.1018 &  & 0.6099 & 0.1565 &  & \cellcolor[HTML]{D0CECE}0.5487 & 0.0519 \\
40/20/10/2 & A & 0.6205 & 0.1003 &  & 0.7377 & 0.2249 &  & 0.5785 & 0.0661 &  & 0.6332 & 0.0900 &  & \cellcolor[HTML]{D0CECE}0.5612 & 0.0661 \\
50/10/10/1 & A & 0.8442 & 0.0971 &  & 0.9234 & 0.2776 &  & 0.7564 & 0.1171 &  & 0.8474 & 0.1088 &  & \cellcolor[HTML]{D0CECE}0.7266 & 0.0939 \\
50/10/10/2 & KW & 0.8873 & 0.1177 &  & 0.9444 & 0.3781 &  & 0.8609 & 0.2057 &  & 0.9095 & 0.1191 &  & \cellcolor[HTML]{D0CECE}0.8288 & 0.2300 \\
50/20/10/1 & KW & 0.5739 & 0.0919 &  & 0.7260 & 0.3458 &  & 0.5829 & 0.1193 &  & 0.5557 & 0.1437 &  & \cellcolor[HTML]{BDD7EE}0.5265 & 0.0728 \\
50/20/10/2 & KW & 0.6237 & 0.1621 &  & 0.8223 & 0.3230 &  & 0.5821 & 0.0711 &  & 0.5957 & 0.1095 &  & \cellcolor[HTML]{BDD7EE}0.4947 & 0.0615 \\
30/10/20/1 & A & 0.7866 & 0.1342 &  & 0.9224 & 0.2362 &  & 0.5752 & 0.0772 &  & 0.7337 & 0.1304 &  & \cellcolor[HTML]{D0CECE}0.5714 & 0.0826 \\
30/10/20/2 & KW & 0.6921 & 0.1251 &  & 0.8991 & 0.2855 &  & 0.5733 & 0.0758 &  & 0.7766 & 0.1526 &  & \cellcolor[HTML]{BDD7EE}\textbf{0.4910} & 0.0663 \\
30/20/20/1 & KW & 0.5509 & 0.0983 &  & 0.5937 & 0.3159 &  & 0.5180 & 0.0708 &  & 0.4756 & 0.0866 &  & \cellcolor[HTML]{D0CECE}0.4485 & 0.0549 \\
30/20/20/2 & KW & 0.6080 & 0.1369 &  & 0.5341 & 0.2132 &  & 0.5295 & 0.0856 &  & 0.5311 & 0.0717 &  & \cellcolor[HTML]{BDD7EE}\textbf{0.4246} & 0.0620 \\
40/10/20/1 & A & 0.8281 & 0.1369 &  & 0.8635 & 0.2682 &  & 0.7605 & 0.0987 &  & 0.8115 & 0.1049 &  & \cellcolor[HTML]{D0CECE}0.7310 & 0.1285 \\
40/10/20/2 & KW & 0.8799 & 0.1429 &  & 0.9173 & 0.2136 &  & 0.8373 & 0.1597 &  & 0.8646 & 0.1488 &  & \cellcolor[HTML]{D0CECE}0.7551 & 0.1365 \\
40/20/20/1 & A & 0.7236 & 0.0910 &  & 0.8042 & 0.2034 &  & \cellcolor[HTML]{D0CECE}0.6443 & 0.0861 &  & 0.6992 & 0.1065 &  & 0.6568 & 0.0920 \\
40/20/20/2 & A & 0.6479 & 0.1080 &  & 0.7267 & 0.2330 &  & 0.5629 & 0.0637 &  & 0.6338 & 0.0869 &  & \cellcolor[HTML]{D0CECE}0.5313 & 0.0818 \\
50/10/20/1 & KW & 0.9315 & 0.1360 &  & 1.1489 & 0.3325 &  & \cellcolor[HTML]{BDD7EE}\textbf{0.8070} & 0.1367 &  & 0.9172 & 0.1132 &  & 0.8570 & 0.1718 \\
50/10/20/2 & A & 0.9184 & 0.1106 &  & 0.8355 & 0.2501 &  & 0.7882 & 0.1230 &  & 0.8949 & 0.0944 &  & \cellcolor[HTML]{D0CECE}0.7833 & 0.1302 \\
50/20/20/1 & A & 0.6921 & 0.1048 &  & 0.7777 & 0.2343 &  & 0.6245 & 0.0679 &  & 0.6731 & 0.1334 &  & \cellcolor[HTML]{D0CECE}0.6140 & 0.0967 \\
50/20/20/2 & A & 0.7817 & 0.1087 &  & 0.8277 & 0.2789 &  & 0.7184 & 0.1110 &  & 0.7399 & 0.0975 &  & \cellcolor[HTML]{D0CECE}0.6763 & 0.1239 \\
30/10/40/1 & A & 0.6146 & 0.0928 &  & 0.8078 & 0.2284 &  & 0.5636 & 0.0908 &  & 0.6541 & 0.0936 &  & \cellcolor[HTML]{D0CECE}0.5377 & 0.1033 \\
30/10/40/2 & KW & 0.8327 & 0.1429 &  & 0.9267 & 0.4573 &  & \cellcolor[HTML]{D0CECE}0.6803 & 0.1318 &  & 0.8237 & 0.1624 &  & 0.7106 & 0.1321 \\
30/20/40/1 & A & 0.6511 & 0.0905 &  & 0.8261 & 0.2427 &  & \cellcolor[HTML]{D0CECE}0.5624 & 0.0714 &  & 0.6940 & 0.0847 &  & 0.5668 & 0.1040 \\
30/20/40/2 & KW & 0.6258 & 0.0931 &  & \cellcolor[HTML]{BDD7EE}\textbf{0.5228} & 0.2466 &  & 0.6227 & 0.0780 &  & 0.6048 & 0.1194 &  & 0.5333 & 0.1241 \\
40/10/40/1 & KW & 0.7928 & 0.1230 &  & 0.9855 & 0.4728 &  & 0.8009 & 0.1691 &  & 0.8536 & 0.2520 &  & \cellcolor[HTML]{D0CECE}0.7451 & 0.1467 \\
40/10/40/2 & KW & 0.9764 & 0.1013 &  & 0.9434 & 0.2664 &  & 0.9314 & 0.1013 &  & 0.9438 & 0.0713 &  & \cellcolor[HTML]{D0CECE}0.7653 & 0.2099 \\
40/20/40/1 & A & 0.8412 & 0.0872 &  & 0.8855 & 0.1983 &  & \cellcolor[HTML]{D0CECE}0.7725 & 0.0777 &  & 0.8491 & 0.0842 &  & 0.8012 & 0.1053 \\
40/20/40/2 & KW & 0.7523 & 0.1216 &  & 0.8667 & 0.3850 &  & \cellcolor[HTML]{BDD7EE}\textbf{0.6327} & 0.0795 &  & 0.6657 & 0.2394 &  & 0.6482 & 0.1492 \\
50/10/40/1 & KW & 0.9508 & 0.1246 &  & 1.0678 & 0.2223 &  & 0.8731 & 0.1045 &  & 0.9504 & 0.0932 &  & \cellcolor[HTML]{BDD7EE}\textbf{0.8245} & 0.1533 \\
50/10/40/2 & KW & 0.9550 & 0.0972 &  & 0.9849 & 0.3697 &  & 0.8978 & 0.1345 &  & 0.8860 & 0.1847 &  & \cellcolor[HTML]{D0CECE}0.8115 & 0.1647 \\
50/20/40/1 & A & 0.8164 & 0.0992 &  & 0.8674 & 0.1941 &  & \cellcolor[HTML]{D0CECE}0.6978 & 0.1003 &  & 0.7956 & 0.0858 &  & 0.7250 & 0.1334 \\
50/20/40/2 & KW & 0.8335 & 0.0912 &  & 0.9774 & 0.2854 &  & 0.8427 & 0.1172 &  & 0.8333 & 0.1226 &  & \cellcolor[HTML]{BDD7EE}\textbf{0.7743} & 0.1443 \\
30/10/60/1 & A & 0.8559 & 0.0976 &  & \cellcolor[HTML]{D0CECE}0.7604 & 0.1764 &  & 0.7605 & 0.1152 &  & 0.8472 & 0.0900 &  & 0.7702 & 0.1085 \\
30/10/60/2 & KW & 0.7789 & 0.1136 &  & \cellcolor[HTML]{D0CECE}0.6548 & 0.3305 &  & 0.6920 & 0.1698 &  & 0.7579 & 0.1347 &  & 0.6572 & 0.2077 \\
30/20/60/1 & KW & 0.6461 & 0.1111 &  & 0.7535 & 0.2920 &  & 0.5599 & 0.1016 &  & 0.6046 & 0.0734 &  & \cellcolor[HTML]{D0CECE}0.5092 & 0.0678 \\
30/20/60/2 & KW & 0.6496 & 0.0770 &  & \cellcolor[HTML]{D0CECE}0.5791 & 0.3243 &  & 0.6093 & 0.0656 &  & 0.6280 & 0.1258 &  & 0.5915 & 0.1575 \\
40/10/60/1 & KW & 0.8705 & 0.0956 &  & 1.0243 & 0.2226 &  & 0.7900 & 0.1766 &  & 0.8971 & 0.1259 &  & \cellcolor[HTML]{BDD7EE}\textbf{0.7838} & 0.1891 \\
40/10/60/2 & KW & 0.8882 & 0.1273 &  & 1.0672 & 0.2353 &  & \cellcolor[HTML]{BDD7EE}\textbf{0.8217} & 0.0989 &  & 0.8930 & 0.0992 &  & 0.8263 & 0.1472 \\
40/20/60/1 & KW & 0.7481 & 0.1313 &  & 0.8326 & 0.4477 &  & \cellcolor[HTML]{D0CECE}0.6887 & 0.1777 &  & 0.8093 & 0.0862 &  & 0.7320 & 0.0869 \\
40/20/60/2 & KW & 0.8597 & 0.1656 &  & 0.8370 & 0.2320 &  & 0.7834 & 0.1456 &  & 0.8203 & 0.1585 &  & \cellcolor[HTML]{D0CECE}0.7561 & 0.1690 \\
50/10/60/1 & A & 0.9455 & 0.0777 &  & 1.0507 & 0.2146 &  & 0.9029 & 0.0509 &  & 0.9458 & 0.0542 &  & \cellcolor[HTML]{BDD7EE}\textbf{0.8055} & 0.1237 \\
50/10/60/2 & KW & 0.9634 & 0.1223 &  & 1.0920 & 0.1976 &  & 0.9338 & 0.0687 &  & 0.9393 & 0.0508 &  & \cellcolor[HTML]{BDD7EE}\textbf{0.8353} & 0.1737 \\
50/20/60/1 & A & 0.8837 & 0.0893 &  & 0.9722 & 0.2007 &  & \cellcolor[HTML]{D0CECE}0.8162 & 0.1304 &  & 0.8414 & 0.0845 &  & 0.8259 & 0.1253 \\
50/20/60/2 & KW & 0.8932 & 0.1281 &  & 0.9357 & 0.4113 &  & 0.7836 & 0.1153 &  & 0.8749 & 0.0726 &  & \cellcolor[HTML]{D0CECE}0.7573 & 0.1060
\\
\bottomrule
\end{tabular}}
\caption{Results of spread metric for the studied MOEAs.\label{tab:spread}}
\end{table}

%\fi

\subsubsection{Consolidated Pareto fronts}

Consolidated Pareto fronts were built for each studied MOEA by applying a dominance check over the non-dominated solutions computed in the 30 independent executions performed for each problem instance. 

\begin{figure*}[!ht]
\centering
    \captionsetup{justification=centering}
    \captionsetup[subfigure]{skip=12pt}
    \begin{subfigure}[b]{0.475\textwidth}
        \centering
        \includegraphics[scale=0.4,trim={0.2cm 0.1cm 0.2cm 0.3cm},clip]{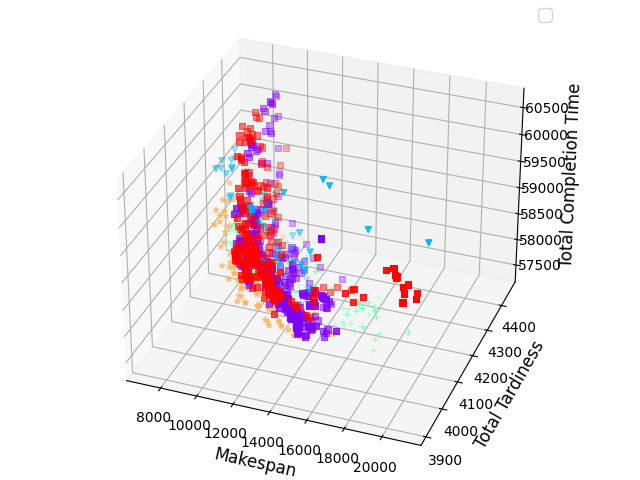}
        \caption{instance 50/20/0/1}
        \label{fig:sample_50/20/0/1}
    \end{subfigure}
    \hfill
    \begin{subfigure}[b]{0.475\textwidth}  
        \centering 
        \includegraphics[scale=0.4,trim={0cm 0cm 0cm 0cm},clip]{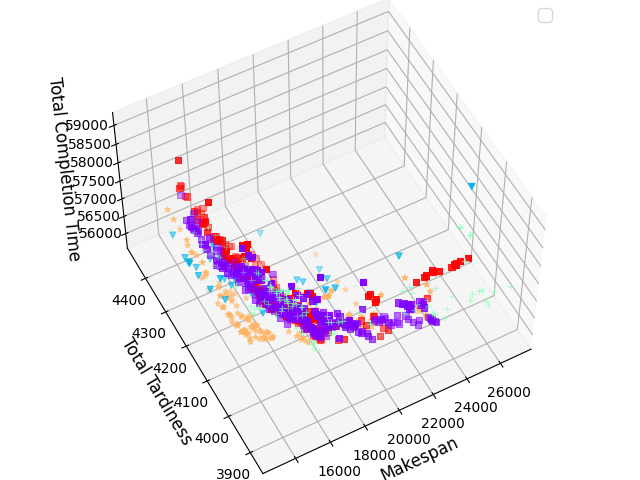}
        \caption{instance 50/20/0/2}
        \label{fig:sample_50/20/0/2}
    \end{subfigure}
    \vskip\baselineskip
    \begin{subfigure}[b]{0.475\textwidth}  
        \centering 
        \includegraphics[scale=0.4,trim={0cm 0cm 0cm 0cm},clip]{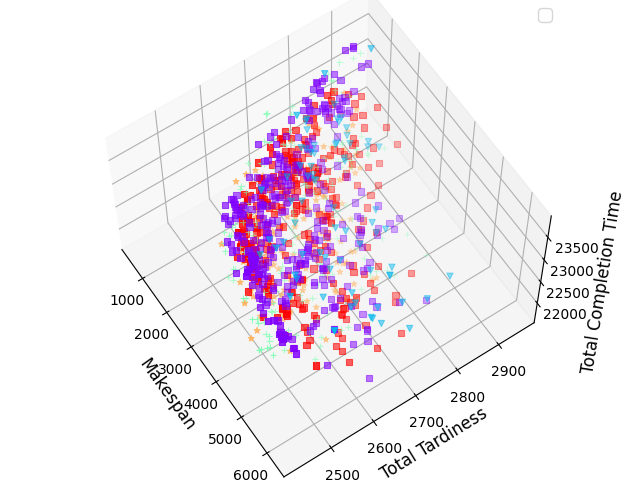}
        \caption{instance 30/20/20/1}
        \label{fig:sample_30/20/20/1}
    \end{subfigure}
    \hfill
    \begin{subfigure}[b]{0.475\textwidth}   
        \centering 
        \includegraphics[scale=0.4,trim={0cm 0cm 0cm 0cm},clip]{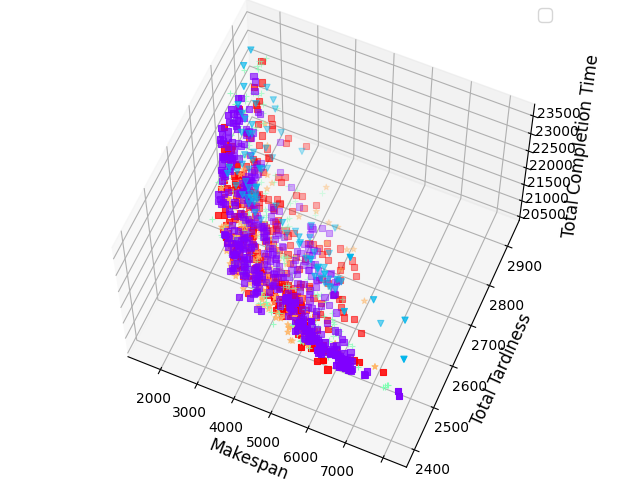}
        \caption{instance 30/20/20/2}
        \label{fig:sample_30/20/20/2}
    \end{subfigure}
    \caption[]
    {Sample Pareto fronts for representative problem instances.\\ 
    \begin{tikzpicture}
        \draw node[name=square, fill=violet] {};
    \end{tikzpicture}
        AGMOEA-II,
    \begin{tikzpicture}
        \draw node[name=square, fill=red] {};
    \end{tikzpicture}
        SPEA2,
    \begin{tikzpicture}
        %\draw node[name=square, fill=orange] {};
        \draw node[name=star, star, star point height=.1cm, minimum size=.4cm, inner sep=0,outer sep=0, fill=orange] {};
    \end{tikzpicture}
        NSGA-III,
    \begin{tikzpicture}
        \draw node[name=square, fill=white, text=cyan, inner sep=0pt]{+};
    \end{tikzpicture}
        NSGA-II,
    \resizebox{0.3cm}{0.3cm}{%
    \begin{tikzpicture}
        \draw node[name=triangle, isosceles triangle, isosceles triangle apex angle=60, draw=none, rotate=270, isosceles triangle stretches, fill=white!50!cyan] {};
    \end{tikzpicture}
    }
        MOEA-D.}
    \label{Fig:sample_FPs}
\end{figure*}

As relevant illustrative cases, Fig.~\ref{Fig:sample_FPs} presents the consolidated Pareto fronts for a set of representative instances, computed from the results of the studied MOEAs. These results are examples of the general trends observed for Pareto fronts computed for the other problem instances. In turn, Table~\ref{tab:consolidated} reports the results of the RHV and spread metrics for the consolidated Pareto fronts.

%\iffalse

\begin{table}[!h]
\setlength{\tabcolsep}{2pt}
\renewcommand{\arraystretch}{0.915}
\adjustbox{max width=0.95\textwidth}{%
\centering
\begin{tabular}{l rr r rr r rr r rr r rr}
\toprule
& \multicolumn{2}{c}{AGEMOEA-II} & \multicolumn{1}{c}{} & \multicolumn{2}{c}{MOEA/D} & \multicolumn{1}{c}{} & \multicolumn{2}{c}{NSGA-II} & \multicolumn{1}{c}{} & \multicolumn{2}{c}{NSGA-III} & \multicolumn{1}{c}{} & \multicolumn{2}{c}{SPEA2} \\   \cmidrule{2-3}\cmidrule{5-6}\cmidrule{8-9}\cmidrule{11-12}\cmidrule{14-15}
Instance & \multicolumn{1}{c}{RHV} & \multicolumn{1}{c}{spread} & \multicolumn{1}{c}{} & \multicolumn{1}{c}{RHV} & \multicolumn{1}{c}{spread} & \multicolumn{1}{c}{} & \multicolumn{1}{c}{RHV} & \multicolumn{1}{c}{spread} & \multicolumn{1}{c}{} & \multicolumn{1}{c}{RHV} & \multicolumn{1}{c}{spread} & \multicolumn{1}{c}{} & \multicolumn{1}{c}{RHV} & \multicolumn{1}{c}{spread} \\ \midrule
30/10/0/1 & \cellcolor[HTML]{D9D9D9}0.9519 & 0.5440 &  & 0.7392 & 0.6802 &  & 0.9000 & 0.4744 &  & 0.9023 & 0.5136 &  & 0.9419 & \cellcolor[HTML]{D9D9D9}0.4665 \\
30/10/0/2 & \cellcolor[HTML]{D9D9D9}0.9588 & 0.5560 &  & 0.6670 & 0.5837 &  & 0.9208 & 0.4416 &  & 0.8654 & 0.4525 &  & 0.9027 & \cellcolor[HTML]{D9D9D9}0.4293 \\
30/20/0/1 & \cellcolor[HTML]{D9D9D9}0.9378 & 0.5859 &  & 0.7535 & 0.5835 &  & 0.9348 & \cellcolor[HTML]{D9D9D9}0.4616 &  & 0.9069 & 0.5188 &  & 0.9204 & 0.4655 \\
30/20/0/2 & 0.9586 & 0.5373 &  & 0.7522 & 0.6371 &  & \cellcolor[HTML]{D9D9D9}0.9638 & 0.4865 &  & 0.9331 & \cellcolor[HTML]{D9D9D9}0.4715 &  & 0.9417 & 0.5279 \\
40/10/0/1 & 0.9005 & 0.8004 &  & 0.6813 & 0.7807 &  & 0.9065 & \cellcolor[HTML]{D9D9D9}0.5344 &  & 0.9102 & 0.7246 &  & \cellcolor[HTML]{D9D9D9}0.9524 & 0.5988 \\
40/10/0/2 & 0.8985 & 0.7309 &  & 0.8014 & 0.8233 &  & 0.9229 & 0.6502 &  & \cellcolor[HTML]{D9D9D9}0.9703 & \cellcolor[HTML]{D9D9D9}0.5258 &  & 0.9426 & 0.5625 \\
40/20/0/1 & \cellcolor[HTML]{D9D9D9}0.9793 & 0.5652 &  & 0.6902 & 0.5389 &  & 0.9336 & 0.4987 &  & 0.9380 & 0.5156 &  & 0.9221 & \cellcolor[HTML]{D9D9D9}0.4837 \\
40/20/0/2 & 0.8668 & 0.6910 &  & 0.6162 & 0.7422 &  & \cellcolor[HTML]{D9D9D9}0.9629 & \cellcolor[HTML]{D9D9D9}0.5027 &  & 0.8964 & 0.5251 &  & 0.8592 & 0.5271 \\
50/10/0/1 & 0.8589 & 0.8395 &  & 0.7153 & 1.1333 &  & 0.8066 & \cellcolor[HTML]{D9D9D9}0.7105 &  & \cellcolor[HTML]{D9D9D9}0.8914 & 1.0291 &  & 0.8269 & 0.7702 \\
50/10/0/2 & 0.8396 & 0.9400 &  & 0.6201 & 0.9237 &  & \cellcolor[HTML]{D9D9D9}0.9111 & \cellcolor[HTML]{D9D9D9}0.6416 &  & 0.8841 & 0.9331 &  & 0.9013 & 0.7814 \\
50/20/0/1 & 0.8771 & 0.5886 &  & 0.7370 & 0.6763 &  & 0.8843 & 0.5207 &  & \cellcolor[HTML]{D9D9D9}0.9433 & 0.5176 &  & 0.8532 & \cellcolor[HTML]{D9D9D9}0.4813 \\
50/20/0/2 & \cellcolor[HTML]{D9D9D9}0.9481 & 0.6193 &  & 0.7439 & 0.7822 &  & 0.8870 & 0.5521 &  & 0.9141 & 0.5987 &  & 0.8905 & \cellcolor[HTML]{D9D9D9}0.5081 \\
30/10/10/1 & \cellcolor[HTML]{D9D9D9}0.9573 & \cellcolor[HTML]{D9D9D9}0.5231 &  & 0.8307 & 0.7798 &  & 0.9551 & 0.5283 &  & 0.9389 & 0.5944 &  & 0.9482 & 0.6100 \\
30/10/10/2 & 0.9195 & 0.5545 &  & 0.8011 & 0.7345 &  & 0.9422 & \cellcolor[HTML]{D9D9D9}0.4331 &  & \cellcolor[HTML]{D9D9D9}0.9675 & 0.6520 &  & 0.9525 & 0.6443 \\
30/20/10/1 & \cellcolor[HTML]{D9D9D9}0.9603 & 0.5462 &  & 0.7928 & 0.6123 &  & 0.9576 & 0.4556 &  & 0.9349 & 0.5921 &  & 0.9271 & \cellcolor[HTML]{D9D9D9}0.4033 \\
30/20/10/2 & 0.9466 & 0.5188 &  & 0.8427 & 0.5290 &  & \cellcolor[HTML]{D9D9D9}0.9703 & \cellcolor[HTML]{D9D9D9}0.5069 &  & 0.9585 & 0.6377 &  & 0.9565 & 0.5327 \\
40/10/10/1 & 0.7903 & 0.7210 &  & 0.6198 & \cellcolor[HTML]{D9D9D9}0.6842 &  & \cellcolor[HTML]{D9D9D9}0.9855 & 0.7670 &  & 0.8589 & 0.9604 &  & 0.8870 & 0.7849 \\
40/10/10/2 & 0.8609 & \cellcolor[HTML]{D9D9D9}0.6261 &  & 0.7052 & 0.9046 &  & \cellcolor[HTML]{D9D9D9}0.9478 & 0.6842 &  & 0.9274 & 0.6981 &  & 0.9317 & 0.6268 \\
40/20/10/1 & \cellcolor[HTML]{D9D9D9}0.9662 & 0.5851 &  & 0.6509 & 0.5798 &  & 0.9088 & 0.4987 &  & 0.9219 & \cellcolor[HTML]{D9D9D9}0.4838 &  & 0.8995 & 0.5487 \\
40/20/10/2 & 0.9146 & 0.5636 &  & 0.6879 & 0.7254 &  & 0.9341 & \cellcolor[HTML]{D9D9D9}0.5166 &  & 0.9314 & 0.5402 &  & \cellcolor[HTML]{D9D9D9}0.9398 & 0.5922 \\
50/10/10/1 & 0.8212 & 0.8255 &  & 0.6851 & 0.9384 &  & 0.8353 & \cellcolor[HTML]{D9D9D9}0.5639 &  & \cellcolor[HTML]{D9D9D9}0.9353 & 0.6974 &  & 0.8419 & 0.7361 \\
50/10/10/2 & 0.8364 & 0.7498 &  & 0.8662 & 0.9874 &  & \cellcolor[HTML]{D9D9D9}0.9618 & 0.6965 &  & 0.9227 & 0.8370 &  & 0.9240 & \cellcolor[HTML]{D9D9D9}0.5972 \\
50/20/10/1 & 0.9340 & 0.5707 &  & 0.6653 & 0.7260 &  & \cellcolor[HTML]{D9D9D9}0.9375 & \cellcolor[HTML]{D9D9D9}0.4506 &  & 0.9148 & 0.5557 &  & 0.9177 & 0.7406 \\
50/20/10/2 & 0.9069 & 0.6045 &  & 0.7838 & 0.6860 &  & 0.9187 & 0.5195 &  & \cellcolor[HTML]{D9D9D9}0.9638 & 0.6459 &  & 0.8760 & \cellcolor[HTML]{D9D9D9}0.4873 \\
30/10/20/1 & 0.9498 & 0.7138 &  & 0.6672 & 0.6491 &  & 0.9677 & 0.5181 &  & 0.9124 & 0.7285 &  & \cellcolor[HTML]{D9D9D9}0.9760 & \cellcolor[HTML]{D9D9D9}0.4781 \\
30/10/20/2 & 0.9327 & \cellcolor[HTML]{D9D9D9}0.3966 &  & 0.7172 & 0.7126 &  & \cellcolor[HTML]{D9D9D9}0.9631 & 0.4974 &  & 0.9027 & 0.5711 &  & 0.9266 & 0.5212 \\
30/20/20/1 & 0.9035 & 0.5090 &  & 0.6683 & 0.6441 &  & 0.9282 & \cellcolor[HTML]{D9D9D9}0.3759 &  & \cellcolor[HTML]{D9D9D9}0.9553 & 0.4484 &  & 0.8792 & 0.4692 \\
30/20/20/2 & 0.9359 & 0.4594 &  & 0.7054 & 0.5763 &  & \cellcolor[HTML]{D9D9D9}0.9704 & 0.4736 &  & 0.9260 & 0.5072 &  & 0.9314 & \cellcolor[HTML]{D9D9D9}0.4234 \\
40/10/20/1 & 0.8378 & 0.9698 &  & 0.7458 & 0.6975 &  & 0.8820 & 0.6077 &  & 0.9163 & 0.6675 &  & \cellcolor[HTML]{D9D9D9}0.9596 & \cellcolor[HTML]{D9D9D9}0.5894 \\
40/10/20/2 & 0.7520 & 0.7832 &  & 0.6613 & 0.8576 &  & \cellcolor[HTML]{D9D9D9}0.8599 & 0.7135 &  & 0.8354 & \cellcolor[HTML]{D9D9D9}0.6069 &  & 0.8262 & 0.7991 \\
40/20/20/1 & 0.8993 & 0.6705 &  & 0.5747 & 0.7599 &  & \cellcolor[HTML]{D9D9D9}0.9232 & 0.5638 &  & 0.8618 & \cellcolor[HTML]{D9D9D9}0.4923 &  & 0.8971 & 0.5148 \\
40/20/20/2 & \cellcolor[HTML]{D9D9D9}0.9608 & 0.6127 &  & 0.6737 & 0.6904 &  & 0.9417 & 0.4956 &  & 0.9132 & 0.6233 &  & 0.9556 & \cellcolor[HTML]{D9D9D9}0.4384 \\
50/10/20/1 & 0.8326 & 0.8539 &  & 0.8144 & 1.0288 &  & 0.9157 & \cellcolor[HTML]{D9D9D9}0.6397 &  & 0.8813 & 0.7216 &  & \cellcolor[HTML]{D9D9D9}0.9815 & 0.7097 \\
50/10/20/2 & 0.7548 & 0.7854 &  & 0.7111 & 0.8814 &  & 0.9041 & \cellcolor[HTML]{D9D9D9}0.6215 &  & 0.8628 & 0.7518 &  & \cellcolor[HTML]{D9D9D9}0.9787 & 0.7918 \\
50/20/20/1 & \cellcolor[HTML]{D9D9D9}0.9147 & 0.5721 &  & 0.7477 & 0.8004 &  & 0.7866 & \cellcolor[HTML]{D9D9D9}0.4934 &  & 0.8974 & 0.6785 &  & 0.8686 & 0.5731 \\
50/20/20/2 & 0.8985 & 0.6696 &  & 0.7389 & 0.6731 &  & \cellcolor[HTML]{D9D9D9}0.9746 & 0.6243 &  & 0.9135 & 0.6710 &  & 0.8915 & \cellcolor[HTML]{D9D9D9}0.5502 \\
30/10/40/1 & 0.9598 & 0.4911 &  & 0.7364 & 0.7727 &  & 0.9455 & 0.4999 &  & 0.9140 & 0.5648 &  & \cellcolor[HTML]{D9D9D9}0.9618 & \cellcolor[HTML]{D9D9D9}0.4420 \\
30/10/40/2 & 0.8605 & 0.8206 &  & 0.6938 & 0.7542 &  & 0.8655 & 0.5841 &  & \cellcolor[HTML]{D9D9D9}0.9625 & \cellcolor[HTML]{D9D9D9}0.5830 &  & 0.9373 & 0.7213 \\
30/20/40/1 & 0.9494 & 0.5488 &  & 0.7679 & 0.9965 &  & \cellcolor[HTML]{D9D9D9}0.9568 & 0.5402 &  & 0.9050 & 0.5623 &  & 0.9545 & \cellcolor[HTML]{D9D9D9}0.5337 \\
30/20/40/2 & \cellcolor[HTML]{D9D9D9}0.9599 & 0.6059 &  & 0.7427 & 0.6144 &  & 0.9169 & \cellcolor[HTML]{D9D9D9}0.4928 &  & 0.9454 & 0.5943 &  & 0.9157 & 0.6117 \\
40/10/40/1 & 0.9051 & 0.6806 &  & 0.7471 & 0.5990 &  & 0.8940 & 0.6001 &  & \cellcolor[HTML]{D9D9D9}0.9811 & \cellcolor[HTML]{D9D9D9}0.5677 &  & 0.9686 & 0.6897 \\
40/10/40/2 & 0.8915 & 0.9430 &  & 0.8425 & 0.8902 &  & \cellcolor[HTML]{D9D9D9}0.9655 & 1.0119 &  & 0.9114 & 0.8372 &  & 0.8597 & \cellcolor[HTML]{D9D9D9}0.7500 \\
40/20/40/1 & 0.8719 & 0.8269 &  & 0.6862 & 0.8605 &  & 0.9219 & \cellcolor[HTML]{D9D9D9}0.5123 &  & \cellcolor[HTML]{D9D9D9}0.9359 & 0.8069 &  & 0.9088 & 0.7349 \\
40/20/40/2 & \cellcolor[HTML]{D9D9D9}0.9746 & 0.6117 &  & 0.7385 & 0.7277 &  & 0.8402 & \cellcolor[HTML]{D9D9D9}0.4761 &  & 0.8858 & 0.6236 &  & 0.8642 & 0.5590 \\
50/10/40/1 & 0.8540 & 0.9502 &  & 0.6986 & 0.8526 &  & \cellcolor[HTML]{D9D9D9}0.9230 & \cellcolor[HTML]{D9D9D9}0.6382 &  & 0.8932 & 0.6415 &  & 0.8351 & 0.9263 \\
50/10/40/2 & 0.8887 & \cellcolor[HTML]{D9D9D9}0.6430 &  & 0.8142 & 0.7938 &  & \cellcolor[HTML]{D9D9D9}0.9552 & 0.7498 &  & 0.9533 & 0.8955 &  & 0.9109 & 0.8533 \\
50/20/40/1 & 0.8448 & 0.6275 &  & 0.7350 & 0.6294 &  & 0.9085 & \cellcolor[HTML]{D9D9D9}0.5239 &  & \cellcolor[HTML]{D9D9D9}0.9809 & 0.7207 &  & 0.8451 & 0.7088 \\
50/20/40/2 & \cellcolor[HTML]{D9D9D9}0.9335 & 0.6369 &  & 0.6295 & 0.7350 &  & 0.8628 & \cellcolor[HTML]{D9D9D9}0.6332 &  & 0.9141 & 0.6416 &  & 0.7878 & 0.8399 \\
30/10/60/1 & 0.8974 & 0.7123 &  & 0.8181 & 0.7905 &  & \cellcolor[HTML]{D9D9D9}0.9820 & \cellcolor[HTML]{D9D9D9}0.5568 &  & 0.9112 & 0.5882 &  & 0.9398 & 0.7439 \\
30/10/60/2 & 0.9327 & 0.5852 &  & 0.8123 & 0.4804 &  & \cellcolor[HTML]{D9D9D9}0.9842 & \cellcolor[HTML]{D9D9D9}0.4622 &  & 0.9462 & 0.6795 &  & 0.9020 & 0.5799 \\
30/20/60/1 & 0.9387 & 0.7357 &  & 0.8427 & 0.7305 &  & 0.9416 & \cellcolor[HTML]{D9D9D9}0.4584 &  & 0.9245 & 0.6729 &  & \cellcolor[HTML]{D9D9D9}0.9425 & 0.6111 \\
30/20/60/2 & 0.9140 & 0.6769 &  & 0.7560 & 0.7674 &  & \cellcolor[HTML]{D9D9D9}0.9601 & \cellcolor[HTML]{D9D9D9}0.4416 &  & 0.9169 & 0.5051 &  & 0.9297 & 0.5022 \\
40/10/60/1 & 0.8132 & 0.8004 &  & 0.6654 & 1.0265 &  & \cellcolor[HTML]{D9D9D9}0.9278 & \cellcolor[HTML]{D9D9D9}0.6031 &  & 0.8009 & 0.9201 &  & 0.9004 & 0.8120 \\
40/10/60/2 & 0.7905 & \cellcolor[HTML]{D9D9D9}0.6618 &  & 0.6552 & 0.7677 &  & 0.8677 & 0.6809 &  & 0.9134 & 0.7023 &  & \cellcolor[HTML]{D9D9D9}0.9718 & 0.8294 \\
40/20/60/1 & 0.8505 & 0.8627 &  & 0.6992 & 0.9027 &  & 0.9179 & \cellcolor[HTML]{D9D9D9}0.5814 &  & 0.8728 & 0.5938 &  & \cellcolor[HTML]{D9D9D9}0.9600 & 0.6759 \\
40/20/60/2 & 0.8713 & 0.6268 &  & 0.8106 & 0.7264 &  & 0.8755 & 0.6228 &  & \cellcolor[HTML]{D9D9D9}0.9091 & \cellcolor[HTML]{D9D9D9}0.5579 &  & 0.8952 & 0.7525 \\
50/10/60/1 & 0.6909 & \cellcolor[HTML]{D9D9D9}0.7008 &  & 0.6962 & 1.1237 &  & 0.8087 & 0.9109 &  & \cellcolor[HTML]{D9D9D9}0.9194 & 1.0184 &  & 0.8031 & 0.9215 \\
50/10/60/2 & \cellcolor[HTML]{D9D9D9}0.9721 & 0.6410 &  & 0.8176 & 1.0557 &  & 0.8876 & 0.8052 &  & 0.8665 & 0.8295 &  & 0.9561 & \cellcolor[HTML]{D9D9D9}0.5443 \\
50/20/60/1 & 0.7725 & 0.9369 &  & 0.6777 & 0.8747 &  & 0.8679 & \cellcolor[HTML]{D9D9D9}0.5550 &  & \cellcolor[HTML]{D9D9D9}0.9402 & 0.7224 &  & 0.8701 & 0.7689 \\
50/20/60/2 & 0.7841 & 0.8288 &  & 0.6562 & 0.9313 &  & 0.7961 & \cellcolor[HTML]{D9D9D9}0.6653 &  & 0.8950 & 0.8047 &  & \cellcolor[HTML]{D9D9D9}0.9021 & 0.8711
\\
\bottomrule
\end{tabular}}
\caption{Consolidated spread and RHV metrics for the studied MOEAs.}
\label{tab:consolidated}
\end{table}
%\fi

\FloatBarrier
\newpage

The consolidated Pareto fronts computed for instance 50/20/0/1, presented in Fig.~\ref{fig:sample_50/20/0/1}, show that NSGA-III was able to obtain solutions that dominate many of the solutions of the other MOEAs. In turn, the Pareto front obtained by SPEA2 is more distributed through the multidimensional objective space. 
These aspects, derived from visual inspection, align with the results of multiobjective optimization metric reported in Table~\ref{tab:consolidated}, in which NSGA-III obtained the best RHV values but SPEA2 obtained the best spread values. 
Fig.~\ref{fig:sample_50/20/0/2} presents the consolidated Pareto fronts computed for instance 50/20/0/2. Results shows that AGEMOEA-II and NSGA-III obtained good results in terms of dominance of the solutions of the other MOEAs. Then, SPEA2 was able to sample a more distributed front, finding solutions near the extremes of the front. Results in Table~\ref{tab:consolidated} confirm this behavior, showing that AGEMOEA-II and NSGA-III obtained the highest results of RHV and SPEA2 obtained the lowest spread values. Fig.~\ref{fig:sample_30/20/20/1} presents the consolidated Pareto fronts computed for instance 30/20/20/1. Results show that SPEA2 and AGEMOEA were able to obtaine distributed fronts, which is in line to the spread results reported in Table~\ref{tab:consolidated}. Additionally, the solutions of NSGA-II, which obtained the highest RHV results in Table~\ref{tab:consolidated}, clearly dominates all other MOEAs. Similarly, Fig.~\ref{fig:sample_30/20/20/2} presents the consolidated Pareto fronts computed for instance 30/20/20/2, showing that SPEA2 and AGEMOEA were able to obtaine very distributed fronts, which is in line to the spread results reported in Table~\ref{tab:consolidated}.

The analysis of the results of multiobjective optimization metrics reported in Table~\ref{tab:consolidated} indicate that regarding RHV, NSGA-II was able to outperform the other MOEAs in 21 instances, NSGA-III and AGEMOEA-II computed the best RHV results in 14 instances, and SPEA2 computed the best RHV results in 11 instances. Regarding spread, NSGA-II was able to outperform the other MOEAs in 28 instances, SPEA2 computed the best results in 17 instances. The other MOEAs computed the best results in few instances, i.e., NSGA-III in 8 instances, AGEMOEA-II in 6 instances, and MOEA/D in just one instance.
Overall, NSGA-II was able to computed the best results in more instances than any other MOEA in both metrics.

\subsubsection{Impact of missing operations over instances} 

\begin{figure*}[!ht]
    {\centering
    \captionsetup{justification=centering}
    \begin{subfigure}[b]{0.475\textwidth}
        \centering
        \includegraphics[scale=0.4,trim={0.2cm 0.1cm 0.2cm 0.3cm},clip]{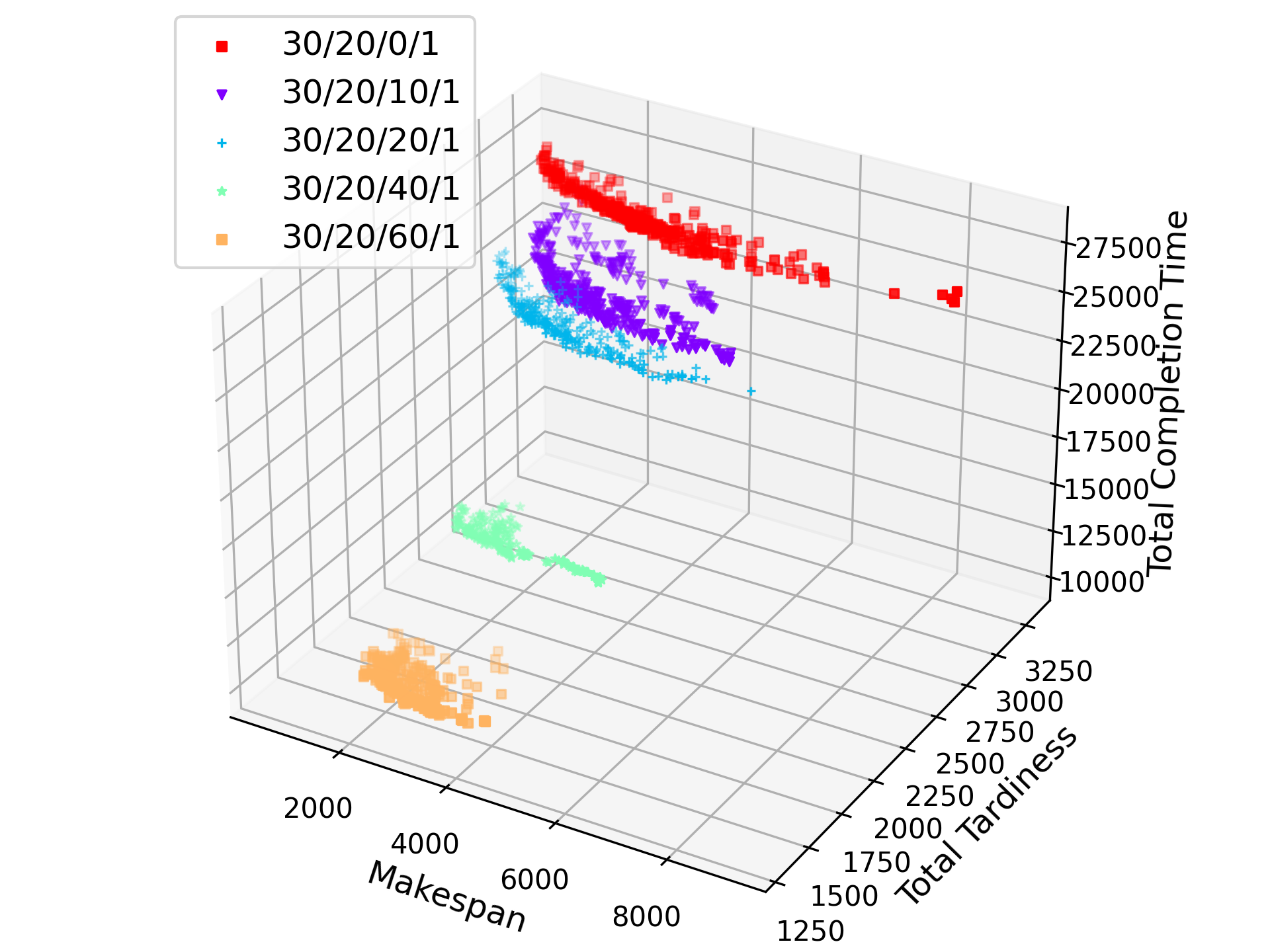}
        \caption[Network2]%
        {{\small 3D Pareto front}}    
        \label{fig:37.50.1}
    \end{subfigure}
    \hfill
    \begin{subfigure}[b]{0.475\textwidth}  
        \centering 
        \includegraphics[scale=0.4,trim={0cm 0cm 0cm 0cm},clip]{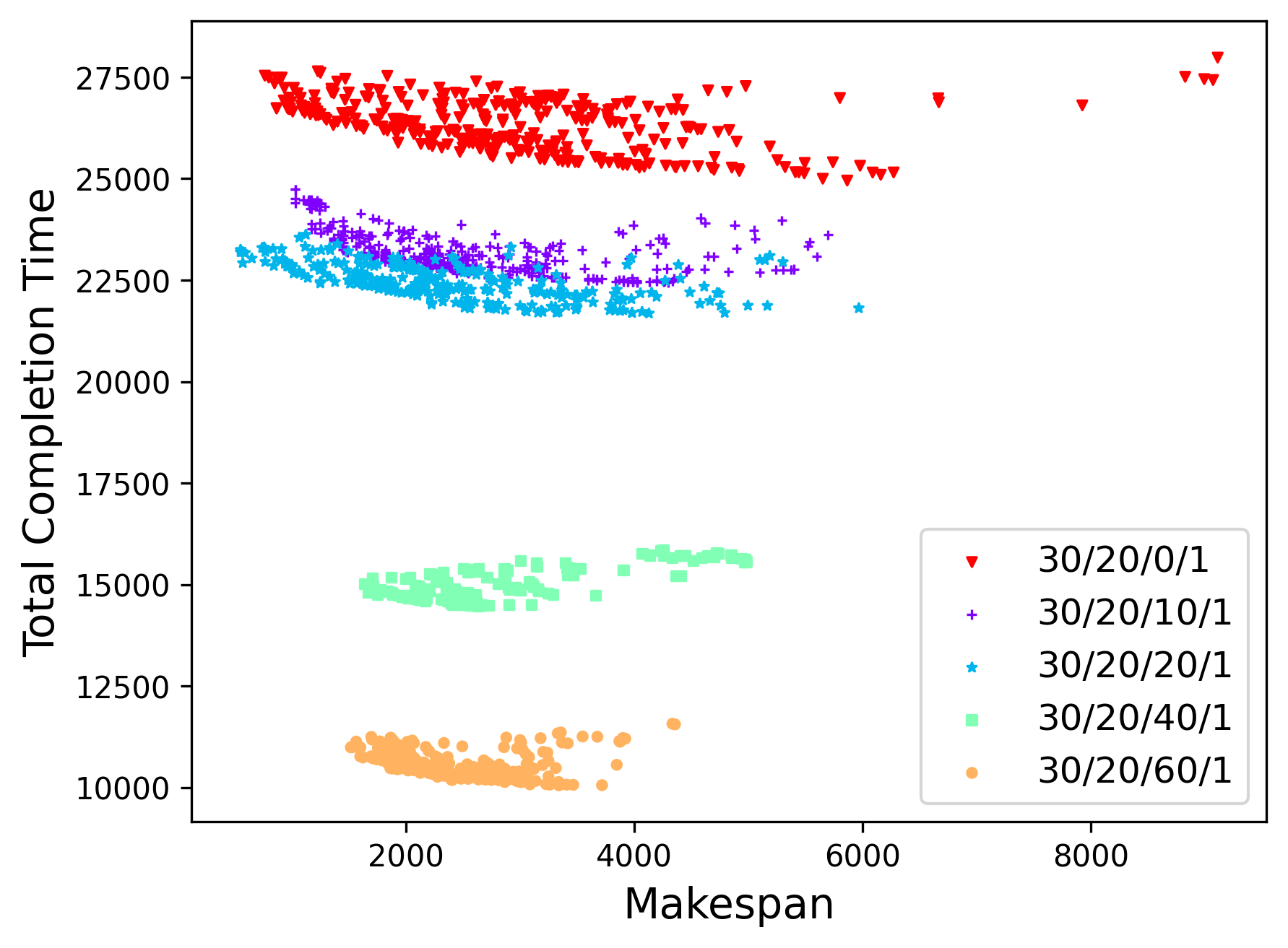}
        \caption[]%
        {{\small 
        % 2D Cut 
        Completion Time/Makespan}}    
        \label{fig:Per_missing_TCT_MKS}
    \end{subfigure}
    \vskip\baselineskip
    \begin{subfigure}[b]{0.475\textwidth}   
        \centering 
        \includegraphics[scale=0.4,trim={0cm 0cm 0cm 0cm},clip]{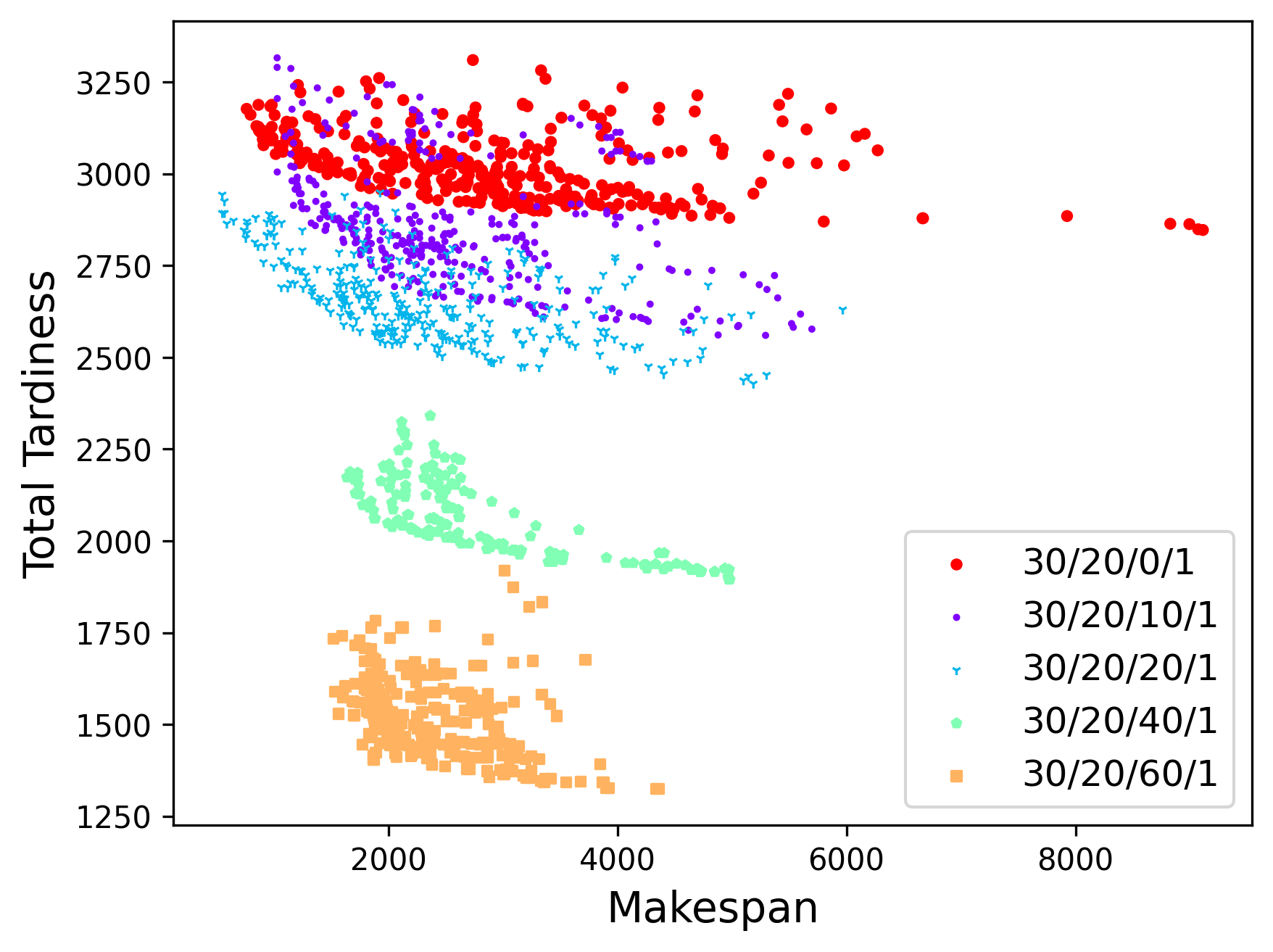}
        \caption[]%
        {{\small 
        % 2D Cut 
        Tardiness/Makespan}}    
        \label{fig:Per_missing_TT_MKS}
    \end{subfigure}
    \hfill
    \begin{subfigure}[b]{0.475\textwidth}   
        \centering 
        \includegraphics[scale=0.4,trim={0cm 0cm 0cm 0cm},clip]{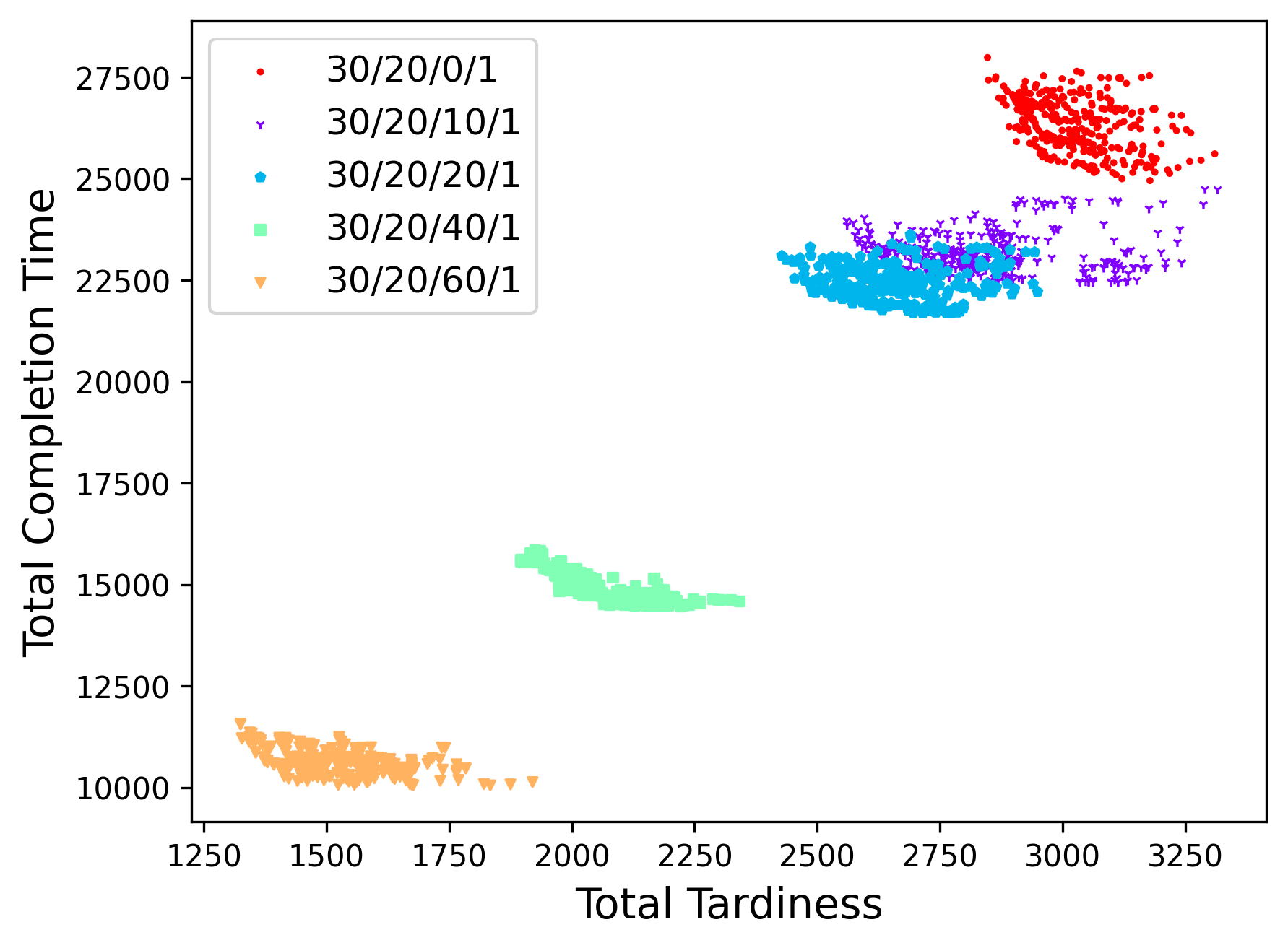}
        \caption[]%
        {{\small 
        % 2D Cut 
        Completion Time/Tardiness}}    
        \label{fig:Per_missing_TCT_TT}
    \end{subfigure}
    \caption{Pareto front of instance 30/20/X/1 with different levels of missing operations (X).\label{fig:Pareto_Fronts_per_missing}}    
    }
\end{figure*}

Another relevant aspect to analyze is whether the different levels of missing operation impact the configuration of the decision-making process, that is, how the different levels of production customization affect the productivity of the system to a greater or lesser extent. In this regard, the relationship between the percentage of missing operation and the values of objective functions was analyzed. Fig.~\ref{fig:Pareto_Fronts_per_missing} presents an analysis of the aforementioned relationship, showing the Pareto fronts obtained considering all the excecutions by any MOEA performed for instance 30/10/$p$/1 with different levels of missing operation (i.e., 0\%, 10\%, 20\%, 40\%, 60\%). The 30/10 dimension was selected for the analysis because the reported values are representative of the general trend observed for the other problem dimensions. Figure~\ref{fig:Pareto_Fronts_per_missing} presents four graphics: a) the 3D front considering the three objective functions, whereas figures b), c), and d) represent the corresponding 2D cuts considering the objective functions of the problem, in pairs. 

%\iffalse
\newpage
%\fi

The Pareto fronts and cuts in Fig.~\ref{fig:Pareto_Fronts_per_missing} show that the percentage of missing operation affects the value of the objective functions. However, not all the objective functions are equally sensitive to the missing operations percentage, then, the impact of customization is dependent on the performance criterion that is considered. The most sensitive objective was Total Tardiness, as Fig.~\ref{fig:Per_missing_TT_MKS} and ~\ref{fig:Per_missing_TCT_TT} depicts. On these plots it is shown that depending on the percentage of missing operation, the value of Total Tardiness varied from 3,250 (with no missing operations, i.e. 0\%) to 1,300 (with 60\% missing operations), what means a decrease on a ratio of over 2.5 just for incorporating missing operation. Also Total Completion time has a similar behaviour, the higher the percentage of missing operation, the lower the objective value. In this case the ratio from maximum largest Total Completion time (27,500) for non-missing operation and the minimum (10,000 and maximal percentage of missing operation), is about 2.75. In counterpart, the makespan objective did not show a very sensitive behavior to the percentage of missing operations, having a similar value regardless the percentage of missing , between 2,000 and 4,000. 

\section{Conclusions and future work\label{sec_conclusion}}

Mass customization, as part of the Smart industry paradigm, seeks to combine the advantages of mass production with those of customization. 
This article studied the flow shop problem with missing operations that arises in shop-floor operations as a consequence of mass customization. Five state-of-the-art MOEAs were applied to simultaneously optimize three traditional metrics of flowshop problems: weighted Total Completion Time, total tardiness and makespan. In the computational analysis over realistic instances, NSGA-II consistently computed the best results, regarding
both convergence and diversity metrics. Moreover, another relevant result is the greater impact of the percentage of missing operations on total tardiness and weighted total completion time, whereas the makespan remained relatively unaffected.

Future work will involve expanding computational experiments by incorporating more instances and higher probabilities of missing operations. Additionally, other Multi-Objective Evolutionary Algorithms (MOEAs) will be included in the analysis to evaluate their performance in this specific problem domain.. Another promising area for future research is studying how personalized production impacts production performance. This aspect has not been thoroughly explored in the literature and is crucial for customized business models, as it provides insights into how production would evolve under these models, directly affecting costs and competitiveness. Therefore, it is important to devote further research efforts to understanding these impacts.

\section*{Acknowledgements} 
This work was partially supported by research projects Red Industria 4.0 (319RT0574, CYTED), PICT-2021-I-INVI-00217 of Agencia I+D+i (Argentina), and PIBAA 0466CO (CONICET).

\section*{Conflict of interest statement} 
On behalf of all authors, the corresponding author states that there is no conflict of interest.

\FloatBarrier

%%===========================================================================================%%
%% If you are submitting to one of the Nature Portfolio journals, using the eJP submission   %%
%% system, please include the references within the manuscript file itself. You may do this  %%
%% by copying the reference list from your .bbl file, paste it into the main manuscript .tex %%
%% file, and delete the associated \verb+\bibliography+ commands.                            %%
%%===========================================================================================%%

\bibliography{Bibliography}% common bib file
%% if required, the content of .bbl file can be included here once bbl is generated
%%\input sn-article.bbl

\end{document}